\documentclass[twoside]{article}

\usepackage[accepted]{aistats2019}
\usepackage{graphicx}       
\usepackage{verbatim}

\usepackage{soul}
\usepackage{tabularx}
\usepackage{float}

\usepackage{hyperref}

\usepackage{color}
\definecolor{redcolor}{rgb}{.7,0.,0.}

\begin{document}

%

%

\twocolumn[

\aistatstitle{A new evaluation framework for topic modeling algorithms based on synthetic corpora}

\aistatsauthor{ Hanyu Shi $^{1}$ \And Martin Gerlach$^{1}$ \And  Isabel Diersen$^1$ \And Doug Downey$^2$ \And Lu\'{i}s A. N. Amaral$^{1, *}$   }

\aistatsaddress{  $^1$Department of Chemical and Biological Engineering, $^2$Department of Electrical Engineering and Computer Science \\
Northwestern University \\ 
Evanston, Illinois 60208, USA\\
$^{*}$amaral@northwestern.edu}
]

%
%

\begin{abstract}
	
Topic models are in widespread use in natural language processing and beyond.
Here, we propose a new framework for the evaluation of probabilistic topic modeling algorithms based on synthetic corpora containing an unambiguously defined ground truth topic structure.
The major innovation of our approach is the ability to quantify the agreement between the planted and inferred topic structures by comparing the assigned topic labels at the level of the tokens.
In experiments, our approach yields novel insights about the relative strengths of topic models as corpus characteristics vary, and the first evidence of an ``undetectable phase'' for topic models when the planted structure is weak.
We also establish the practical relevance of the insights gained for synthetic corpora by predicting the performance of topic modeling algorithms in classification tasks in real-world corpora.

\end{abstract}

\section{Introduction}

Topic modeling is a powerful natural language processing tool for the unsupervised inference of the latent topics of a collection of texts~\cite{Blei2012, Crain2012}.
A variety of topic modeling algorithms have been proposed to cope with a broad set of technical challenges and diverse types of written documents~\cite{Blei2003, Griffiths2004,Blei2007,Buntine2014, Lancichinetti2015}.
Due to the large number of topic models in the literature and their widespread use, it is crucial to benchmark available algorithms.
The need for such approaches is exacerbated by the increase of topic modeling applications in computational social science, where the purpose of the models is not to predict documents (in which case
held-out likelihood would suffice) but instead to help understand the corpus, which requires an evaluation
of the inferred topics themselves~\cite{BoydGraeber2017}.

Our analysis is grounded on the assumption that a hidden topic structure exists in the texts (i.e. a latent variable leading to deviations from the random usage of words). 
Under this assumption, a topic modeling algorithm can be viewed as an instrument for the measurement of the hidden structures.
Crucial to measurement is the existence of a standard that provides ground truth~\cite{Bandalos2018, allen2001}.
For example, the use of synthetic datasets has become standard in order to probe machine learning algorithms in fields such as clustering~\cite{Jain2010} or community detection~\cite{Lancichinetti2008}.

\begin{figure*}[h]
	\begin{center}
		\includegraphics[width=1.5\columnwidth]{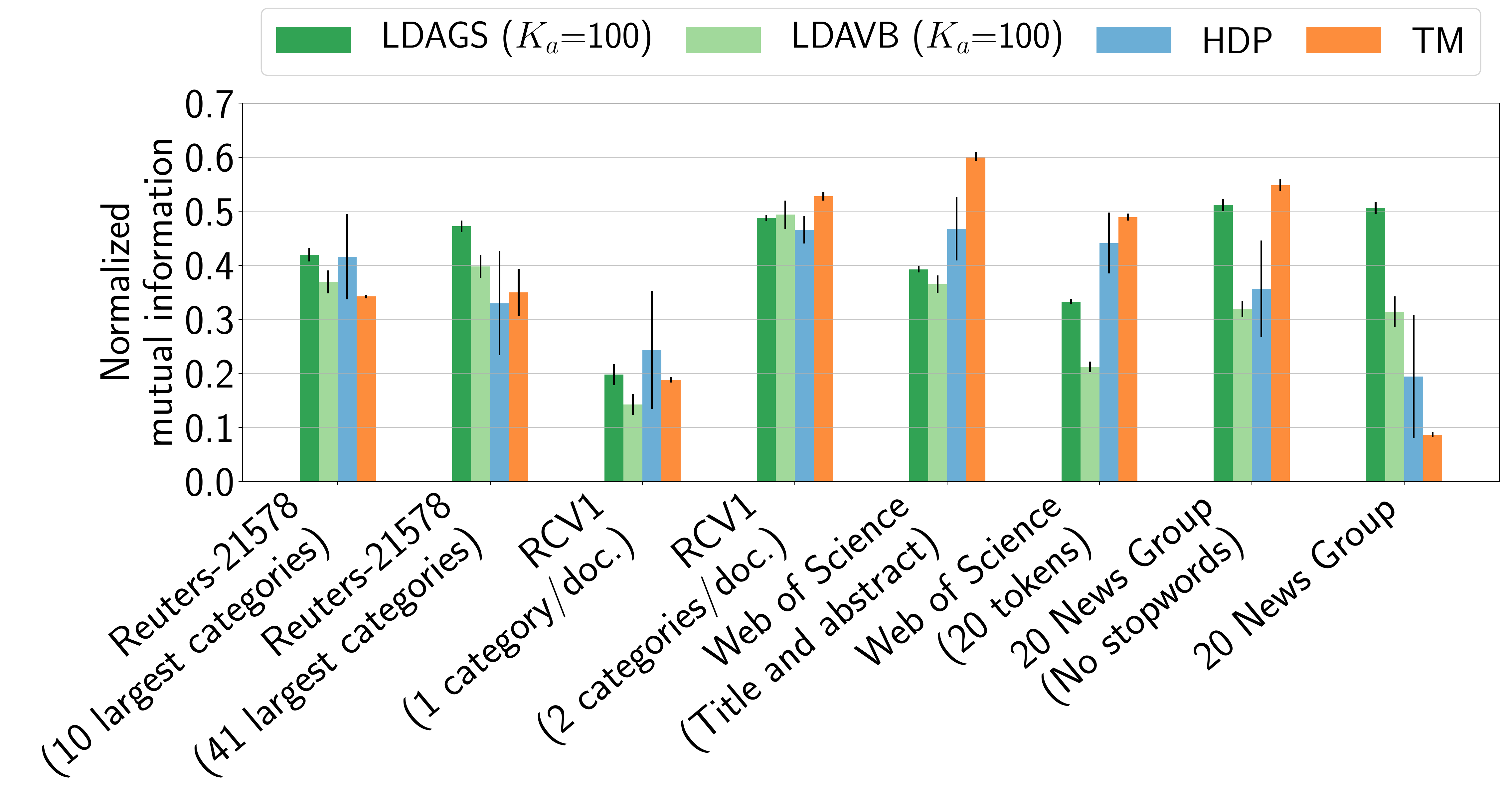}
		\caption{
			\textbf{Performance of topic models is inconsistent across diverse real-world corpora.}
			Normalized mutual information of four topic modeling algorithms in unsupervised document classification for 8 real-world corpora. See \textit{Supplementary Material}, Secs.~S1, S2, and S4 
			for details on the corpora (and the pre-processing steps), the topic modeling algorithms,  and the comparison metric, respectively.
		}
		\label{fig_compare_real}
	\end{center}
\end{figure*}

Currently employed evaluation methods for topic models are often subjective, and can lack theoretical justification.
Indeed, the debate is ongoing as to which evaluation method is best~\cite{Wallach2009a,Chang2009,Roder2015}. 
From a practical perspective, the relative performance of topic modeling algorithms varies substantially across different corpora with different characteristics (see e.g. Fig.~\ref{fig_compare_real}, which compares several topic modeling algorithms on classification tasks). 
While we would expect that certain algorithms or settings are better suited to particular document characteristics (e.g., corpus size, document length, number of topics, burstiness, etc.), it remains unclear how such properties affect the performance of topic modeling algorithms, beyond a certain measure of machine learning ``folklore''~\cite{Tang2014}.

In this work, we present a new framework for topic model evaluation relying on generating a synthetic corpus containing an unambiguous ground truth.
First, we propose a novel way to generate synthetic corpora that generalizes upon previous approaches.
Our approach allows us to isolate the impact of various corpus characteristics, such as size, number of topics, the signal-to-noise ratio, burstiness, or fraction of stopwords, which in real-world corpora are either unknown or impossible to tone.
Second, we propose a new evaluation metric based on the normalized mutual information that compares the agreement between planted and inferred topics on the level of individual word tokens.
Our approach yields an absolute measure of topic modeling accuracy, eliminating the need for post-inference heuristics such as ``topic matching'' \cite{Lancichinetti2015}.
While synthetic ground truth has been used for topic model evaluation in the past, ours is the first framework for evaluating how well topic modeling algorithms perform the key task of inferring per-token topic assignments.
Altogether, the formalization of synthetic corpora allows us to probe more accurately the ability of different topic modeling algorithms to resolve a wide range of topic structures, beyond simplistic assumptions of LDA.
We present experiments showing how different popular topic modeling algorithms fare as these characteristics change, for one type of synthetic corpus. 
We show how our measurement framework leads to new insights, including evidence of an ``undetectable region'' for sufficiently weak topic structures, or how the choice of hyperparameters can bias the inference result.
Finally, we show that our approach is predictive of the performance of topic modeling algorithms in classification tasks in real-world corpora.

\section{Background}

A popular approach for evaluating topic models is to inspect their output manually~\cite{Murakami2017}, but this approach is expensive and subjective. 
The most common quantitative approaches to evaluate topic modeling algorithms rely on intrinsic evaluation methods, such as held-out likelihood~\cite{Wallach2009a}, and topic coherence~\cite{Newman2010, Mimno2011}, or on extrinsic tasks such as document classification~\cite{Lu2011,Xie2013} and information retrieval~\cite{Manning2008,Wei2006}.
However, these approaches allow only for limited insights into why topic modeling algorithms fail or succeed.
For example, perplexity and topic coherence can only provide a relative measure of performance: how well does a topic model do \textit{in relation} to another model?
In contrast, extrinsic evaluation tasks allow for the formulation of absolute measures based on the prediction of document metadata, often considered as ``ground truth'' labels in the literature.
However, extrinsic evaluation approaches, and the latter identification in particular, are also problematic because:
({\it i}) manual labeling is subjective and error prone; 
({\it ii}) they evaluate the topic structure only indirectly (e.g. via the fraction of correctly classified documents); and 
({\it iii}) they implicitly assume that the manually generated labels are truly encoded in the topic structure of the documents.
The latter assumption has been shown to be surprisingly unsupported in other domains~\cite{Hric2014,Peel2016}.

It has been recently shown that topic modeling can be formally mapped to the problem of community detection in networks~\cite{Karrer2011,Gerlach2017}.
The formulation of benchmark corpora pursued here follows the idea of benchmark graphs in community detection.     
There, the basic approach is to build synthetic networks with known (planted) community structure and evaluate an algorithm by comparing the overlap between the planted and the inferred community structures \cite{Lancichinetti2008, Girvan2002, Danon2005, Sales-Pardo15224, Sawardecker2009, Lancichinetti2009, Guimera2007}.
This approach allowed researchers to gain new insights into community detection algorithms such as  ({\it i}) the spurious appearance of large values of modularity in random networks~\cite{Guimera2004}; ({\it ii}) the existence of a resolution limit concerning the minimum size of the groups that can be inferred~\cite{Fortunato2007}; or ({\it iii}) the existence of an undetectable phase in which no algorithm is able to infer a structure~\cite{Decelle2011}.

The use of synthetic corpora has appeared sporadically in the context of topic modeling (see Supplementary Materials, Table~S3). 
In most cases, the synthetic data comes from the generative process of LDA and is tested only on intrinsic evaluation methods such as held-out likelihood~\cite{Wallach2009a} or topic coherence~\cite{AlSumait2009}.
Comparison between planted and inferred structure is usually done by visual inspection~\cite{Griffiths2004,Andrzejewski2009}, focuses only on either the word-topic or topic-document distribution requiring ``matching of topics''~\cite{Taddy2012,Arora2013,Lancichinetti2015}, or evaluate very specific hypothesis of the fitted model (such as the independence of words and documents in individual topics~\cite{Mimno2011}).
Our work formalizes and generalizes these ideas: 
\textit{(i)} by developing a framework to investigate a wide range of topical structures and including a number of realistic features that might be of interest to practitioners;
and \textit{(ii)} proposing a measure that compares the planted and inferred structure (i.e. the topic labels) on the level of individual word tokens.

\section{Evaluating topic modeling algorithms using synthetic corpora}

Our approach to comparing the performance of topic modeling algorithms using synthetic corpora consists of two main steps (Fig.~\ref{fig_token_label})~\footnote{Code to generate synthetic corpora is available at:
\tiny{\url{https://github.com/amarallab/synthetic_benchmark_topic_model}}}.
First, we generate a synthetic benchmark with a planted ground truth structure; and second, we quantify the overlap between the planted and inferred structures.

\subsection{Generating synthetic corpora}
\label{sec.gen.syn.corpus}

\begin{figure*}[h]
	\begin{center}
		\centerline{\includegraphics[width=1.8\columnwidth]{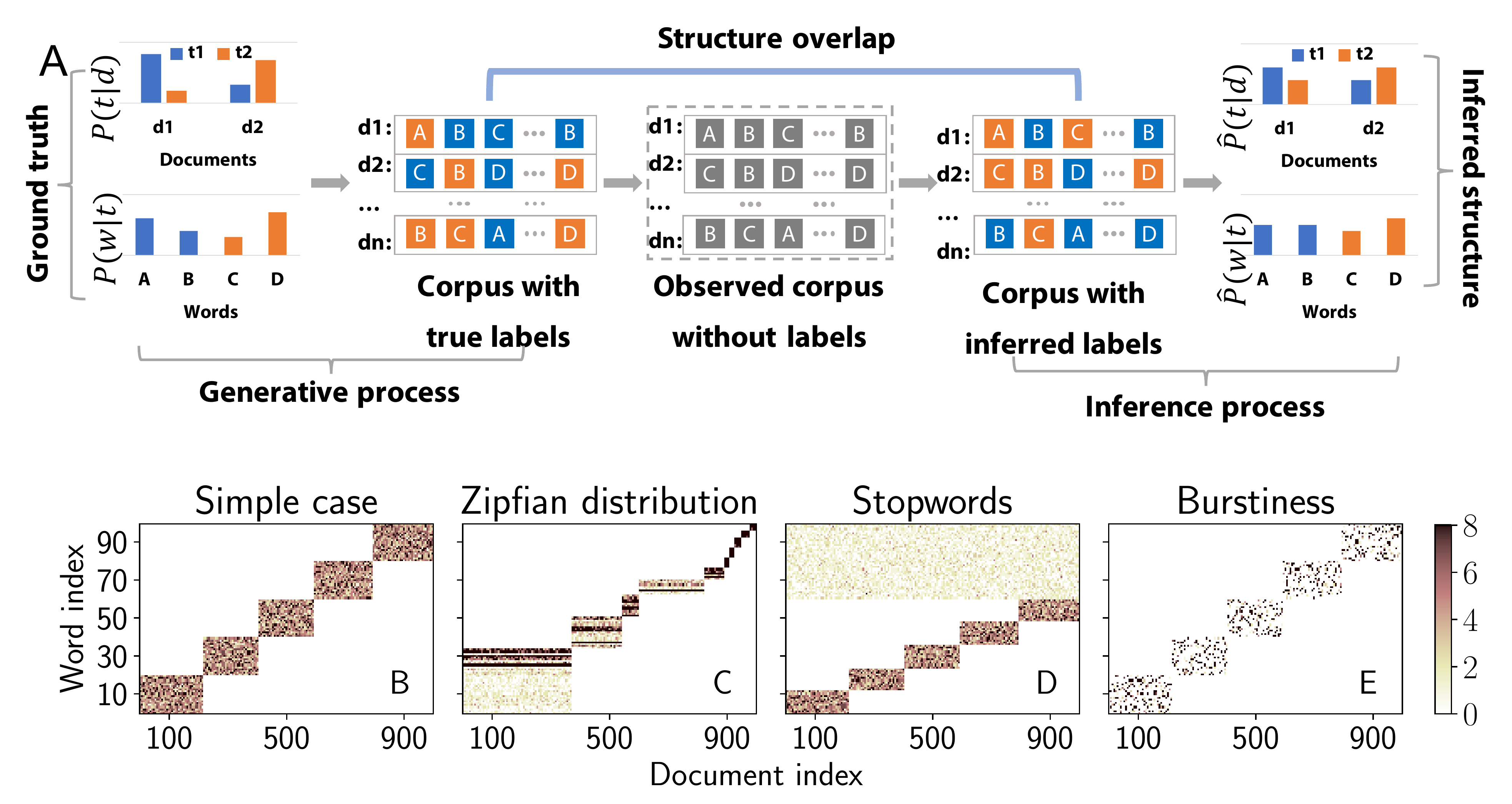}}
		\caption{
			\textbf{Proposed framework for the evaluation of topic models based on synthetic corpora.} 
			\textbf{(A)} Evaluation framework.
			\textbf{(B-E)} Examples of synthetic corpora with different statistical features observed in real-world corpora showing the number of occurrences of each word in each document with $D=1000$ and $V=100$. 
		}
		\label{fig_token_label}
	\end{center}
	\vspace{-.1in}
\end{figure*}

Our approach to generating synthetic corpora with a planted structure is based on the formulation of the generative process employed by probabilistic topic models~\cite{Blei2012, Crain2012}.
Consider a corpus of $d=1,\ldots,D$ documents each with length $m_d$ (and $N=\sum_d m_d$ words in total) generated from $K$ topics and $V$ unique words defining the vocabulary $\mathcal{V}$.
The statistical characteristics of the corpus are determined by two sets of conditional probabilities:
$P(t | d)$, indicating the probability of topic $t$ within document $d$;
and  $P(w | t)$, indicating the probability with which word $w$ is used by topic $t$.
Specifically, for each token $w(i_d)$, defined as the word at position $i_d = 1,\ldots,m_d$ in document $d$, we first draw a topic $z(i_d)=t$ with probability $P(t | d)$ and then a word $w(i_d)=w$ is chosen with probability $P(w | t=z(i_d))$.
Typically, one makes assumptions about these probabilities in the form of prior distributions. For example, in the case of Latent Dirichlet allocation (LDA), it is assumed that $P(t | d)$ and $P(w | t)$ are drawn from Dirichlet distributions with hyperparameters $\alpha$ and $\beta$, respectively.
Given an observed corpus, the aim in topic modeling is then to determine the most likely distributions $\hat P(t | d)$ and $\hat P(w | t)$ by inferring the latent topic variables $\hat z(i_d)$ (Fig.~\ref{fig_token_label}A).

Here, we take the inverse approach by \textit{a priori} fixing the distributions $P(t | d)$ and $P(w | t)$ and using the generative process to produce a synthetic corpus.
Formally, our generation process includes the following steps.

First, we assign each word $w \in \mathcal{V}$ from the vocabulary to either the  stopwords set $\mathcal{V}_S$ ($V_S \equiv \lvert \mathcal{V}_S \rvert$) or topical word set $\mathcal{V}_T$ ($V_T \equiv \lvert \mathcal{V}_T \rvert$) such that $V=V_S + V_T$.

Second, we fix the global word distribution $P(w)$ ($\sum_w P(w)=1$ ) and the number of topical words assigned to each topic $V_t$ ($\sum_t V_t = V_T$).
Here, we consider a uniform or power-law functional form for their distributions.

Third, we assume that each word $w$ belongs uniquely to one topic $t$ denoted by $t_w$ assigned randomly (such that we have $V_t$ words in topic $t$).
This assignment determines the topic distribution $P(t)$ over the entire corpus
\begin{equation}
P(t) = \frac{{\sum\limits_{w \in \mathcal{V}_T} {{\delta _{{t_w},t}} \cdot P(w)} }}
{{\sum\limits_{w \in \mathcal{V}_T} {P(w)} }},
\end{equation}
where $\delta_{i,j}$ is Kronecker delta function, i.e., $\delta_{i,j}=1$ only if $i=j$. 
Assuming that each document $d$ belongs uniquely to one topic denoted by $t_d$ which is randomly assigned with probability $P(t)$.

Fourth, we define the word-topic distribution $P(w|t)$ with structure parameter $c_w$ as
\begin{equation}
\begin{array}{l}
P(w|t) = \\
\left\{ 
\begin{split}
c_w \; \delta_{t_w,t}\frac{P(w)}{P(t)} + (1 - c_w) \; P(w),  & \;  \text{if } w \in \mathcal{V}_T\\
P(w),  &  \; \text{if } w \in \mathcal{V}_S
\end{split} \right.
\end{array} .
\end{equation}
While the topical words ($w \in \mathcal{V}_T$) are characterized by a linear combination of a structured term and a random, unstructured term, the stopwords ($w \in \mathcal{V}_S$) appear randomly in all topics.
Similarly, we define the topic-document distribution $P(t|d)$ with structure parameter $c_d$ as
\begin{equation}
P(t|d) = {c_d} \; {\delta _{t_d,t}} + (1 - {c_d}) \; P(t),
\end{equation}
where the first term is the structured part and the second is the random, unstructured part.

The resulting synthetic corpus contains a fully known planted structure since we know the topic label $z(i_d)$ of each individual token $w(i_d)$ (Fig.~\ref{fig_token_label}A).
The general formulation not only allows us to investigate a wide range of topical structures, but also to incorporate statistical laws observed in real-world corpora~\cite{Altmann2016}, such as Zipfian word-frequency distribution, stopwords, or burstiness (Fig.~\ref{fig_token_label}B-E), see \textit{Supplementary Material} Sec.~S5. 

\subsection{Comparing planted and inferred structure}
Typically, the results of topic modeling algorithms are evaluated either at the level of the topic-document distribution $P(t | d)$ in applications such as document classification, or at the level of the word-topic distribution $P(w | t)$ to judge the topic quality such as in topic coherence \cite{Bhatia2017}. 
Here, we propose a new approach by quantifying the overlap between the planted and the inferred structure based on the comparison of the topic labels of each individual token.

Specifically, for each token $w(i_d)$ we record the planted topic label as $z^{\text{pl}}(i_d)$ and the inferred topic label as $z^{\text{inf}}(i_d)$ 
and construct a confusion matrix $p_{t,t'}$ , which counts the fraction of tokens having a planted topic label $t$ and an inferred topic label $t'$
\begin{equation}
\label{eq.ptt}
p_{t,t'} = \frac{1}{N} \cdot \sum_{d=1}^{D} \sum_{i_d=1}^{m_d} \delta_{ z^{\text{pl}}(i_d),t } \cdot \delta_{  z^{\text{inf}}(i_d),t'  }.
\end{equation}
From this we calculate the normalized mutual information, $\hat{I}$, a commonly used metric to quantify the overlap between different partitions~\cite{Danon2005} defined as:
\begin{equation}
\label{eq.nmi}
\hat I = \frac{ 2I }{ H + H' },
\end{equation}
where $I$ is the mutual information and $H$ (and $H'$) are the respective entropies
\begin{equation}
\begin{alignedat}{3}
I & = \sum_{t} \sum_{t'} p_{t,t'} \log \frac{p_{t,t'}}{p_t p_{t'}},   \\
H & = -\sum_t p_t \log p_t, \,\,\,  H' = -\sum_{t'} p_{t'} \log p_{t'}.
\end{alignedat}
\end{equation}
We thus obtain a measure between $\hat{I}=0$ indicating no overlap, and $\hat{I}=1$ indicating perfect overlap. Note that $\hat{I}$  takes into account that the number of  topics in the inference results does not have to match the number of planted classes (Fig.~S1). 
The major advantage of the NMI is its easy interpretability:  it quantifies the average amount of information one gains about the planted label of a token upon learning its inferred topic label.
Furthermore, $\hat I$ is invariant with respect to permutation of the topic labels;
thus we avoid the issue of finding the ``best match'' between planted and inferred topics, typically addressed by non-trivial heuristic approaches~\cite{Lancichinetti2015} (See ~\cite{Fortunato2010} for advantages of $\hat{I}$ over other measures, such as Jaccard index).

This measure is related to the Variation of Information proposed in~\cite{Schofield_2016}, i.e. $VOI = const. \times (1-\hat{I})$; however, while \cite{Schofield_2016} compare different outputs of a topic modeling algorithm under different pre-processing steps, here we use the measure to compare the planted ground truth against the output of the topic modeling algorithm.

\section{Results}

We report three different experiments that illustrate how synthetic corpora can yield new insights on topic modeling algorithms.
As a representative sample, we evaluate four topic modeling algorithms on these corpora: LDA using Gibbs sampling (LDAGS)~\cite{Griffiths2004,mccallum2002mallet}, LDA using variational inference (LDAVB)~\cite{Blei2003,Rehurek10}, Hierarchical Dirichlet Processes (HDP)~\cite{Teh2006,hdp2017}, and TopicMapping (TM)~\cite{Lancichinetti2015,tm2017} (see \textit{Supplementary Material} Sec.~S2 
for details) using default parameter settings of the corresponding implementations unless stated otherwise.

\subsection{Degree of structure}

Our first experiment evaluates how modeling accuracy varies with the degree of topic structure in the synthetic corpus.
Here, we consider a simple version of the synthetic corpus described in Sec.~\ref{sec.gen.syn.corpus} with a single parameter for the degree of structure $c=c_w=c_d$ such that we can vary between a trivial ($c=1$) and an impossible ($c = 0$) inference problem (as shown in Fig.~S2). 
More specifically, a smaller value of $c$ corresponds to a higher level of noise in the synthetic corpus.
In addition, we fix that  there are no stopwords ($V_s = 0$), and that the global word-distribution and the topic-size distribution are uniform; $P(w)=1/V$ and $V_t = V/K$.   

\begin{figure*}[h]
	\begin{center}
		\includegraphics[width= 1.8\columnwidth]{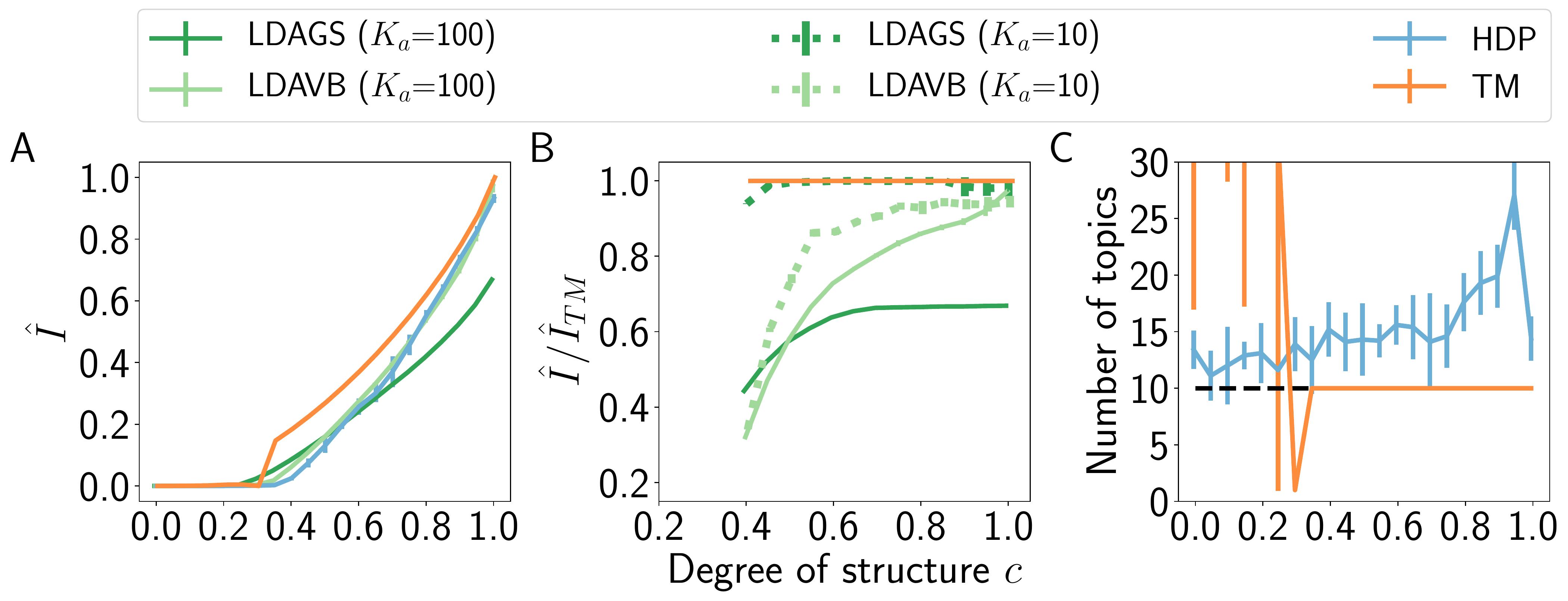}
		\caption{
			\textbf{Performance of topic modeling algorithms in synthetic corpora with varying degree of structure.}
			\textbf{(A)} Normalized mutual information, $\hat I$, between planted and inferred structures for different topic modeling algorithms as a function of the structure parameter $c$.
			\textbf{(B)} Relative performance of different topic modeling algorithms against TopicMapping, the best performing algorithm in \textbf{A}. 
			\textbf{(C)} Number of inferred topics for non-parametric topic modeling algorithms. 
			Synthetic corpora were generated with $K=10$ topics, $D=10^4$ documents, document length $m=100$, and vocabulary size $V=10^3$.
			The lines (error bars) denote averages (one standard deviation) estimated from $10$ realizations.
		}
		\label{fig_token_label_results}
	\end{center}
\vspace{-.1in}
\end{figure*}

In Fig.~\ref{fig_token_label_results} we compare the overlap between planted and inferred structure as a function of $c$ for synthetic corpora with $K=10$ planted topics.

In general, the performance of all algorithms increases non-linearly with the degree of structure $c$ (Fig.~\ref{fig_token_label_results}A). 
We observe substantial differences between algorithms for both the mean (identifying TM as a systematically more accurate algorithm) and the standard deviation (identifying HDP as a systematically less reproducible algorithm).
We also observe a region ( $c<c^*$ with $c^* \approx 0.3$), where none of algorithms are able to recover any structure ($\hat{I}=0$) despite the fact that the synthetic corpus contains some small degree of structure ($c>0$).
The latter suggests the existence of an ``undetectable phase", a phenomenon recently reported in the context of community detection~\cite{Decelle2011}.

For the LDA algorithms in Fig.~\ref{fig_token_label_results}A, we assume the number of topics (a parameter which has to be specified \textit{a priori} in LDA) is $K_a=100$, which is a common choice for real-world corpora in the literature~\cite{Wallach2009a, Wei2006, Aletras2017, Steyvers2007}.
Not surprisingly, in Fig.~\ref{fig_token_label_results}B we observe a substantial improvement in performance when considering the unlikely case of guessing the correct number of topics ($K_a=K=10$).
We find that the performance of LDA algorithms is typically reduced by choosing both too many  or too few topics highlighting how uninformed modeling assumptions can strongly affect performance (Fig.~S3). 

We further investigate how accurately non-parametric topic models such as HDP and TopicMapping can infer the number of topics (Fig.~\ref{fig_token_label_results}C).
TopicMapping finds the correct number of topics even for only moderately structured corpora, but it completely fails for very unstructured corpora by overfitting the data reflecting the intrinsic difficulty when the signal-to-noise ratio is low.
In contrast, HDP tends to overestimate the number of topics in this experiment, even more so as the degree of structure becomes large.  
This suggests that the model is arbitrarily splitting ground truth topics into distinct topics, a hypothesis that is corroborated by the relatively low reproducibility of the method (in terms of the average overlap between two different inferred solutions on the same data, as shown in Fig.~S4). 
Thus, in this experiment we do not find the number of topics inferred by HDP to be reliable.

\subsection{Impact of LDA-implementation and hyperparameter values}

Despite the considerable advances in our understanding of LDA since its original formulation~\cite{Blei2003}, we still lack a systematic understanding of the impact of different approximation techniques on the performance~\cite{Zhang2016}.
While some groups have investigated the advantages of Collapsed Variational Bayes over mean-field Variational Bayes~\cite{Mukherjee2009} or the effect of hyperparameter choice~\cite{Asuncion2009,Wallach2009b},
to our knowledge there have been no systematic studies exploring the inferred solutions in terms of the corresponding topic distributions and how they depend on the hyperparameters or inference algorithms.

\begin{figure*}[h]
	\begin{center}
		\centerline{\includegraphics[width=1.8\columnwidth]{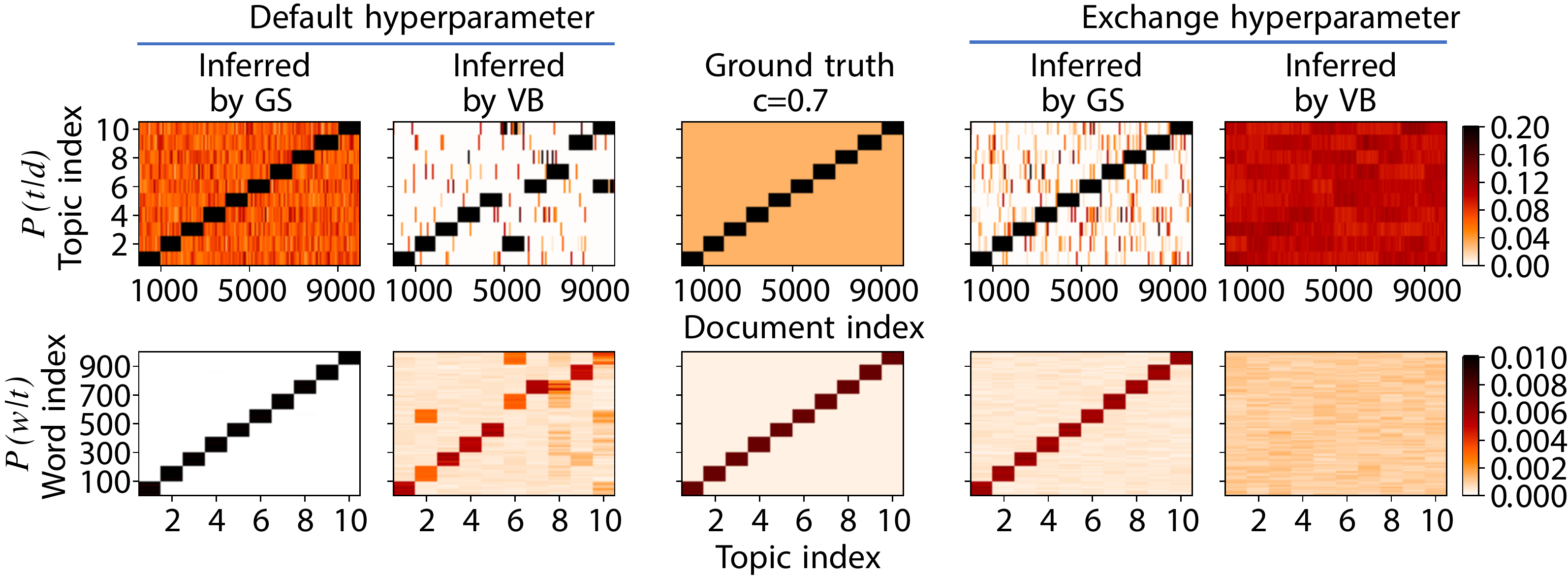}}
		\caption{ 
			\textbf{Default hyperparameter settings bias the inferred topic structure of different LDA implementations.} 
			Comparison of topic distributions $P(t|d)$ (top row) and $P(w|t)$ (bottom row) from the planted and inferred structure from LDAGS and LDAVB using two different sets of hyperparameters: original defaults as defined in each implementation (left panels) and defaults from the other implementation, respectively (right panels).
			Ground truth is displayed in the middle column.
			Same parameters as in Fig.~\ref{fig_token_label_results} fixing $c=0.7$ and using $K_a=10$.
		}
		\label{fig_appx_effect_hyperparameters_ka10}
	\end{center}
\end{figure*}

In order to understand the differences between the Variational Bayes (VB) and Gibbs Sampling (GS) implementations of LDA observed in Fig.~\ref{fig_token_label_results}, we investigate in detail the planted and inferred $P(t | d)$ and $P(w | t)$ for both algorithms (Fig.~\ref{fig_appx_effect_hyperparameters_ka10}).
We find that neither can accurately infer the ground truth topic distributions endowed with a mixed structure in both $P(t | d)$ and $P(w | t)$.
With default hyperparameters, the GS implementation infers a pure word-topic distribution and places the fluctuations almost exclusively on $P(t | d)$  (Fig.~\ref{fig_appx_effect_hyperparameters_ka10}, 1st column).
In contrast, the VB implementation infers a pure topic-document distribution and places the fluctuation mainly on $P(w | t)$ (Fig.~\ref{fig_appx_effect_hyperparameters_ka10}, 2nd column).
However, these differences can be explained, in part, by different default values for the hyperparameters.
Assuming the correct number of topics ($K_a=10$) and using the same hyperparameters (default values from VB implementation)
for both the GS and the VB inference, we obtain almost identical results from the two LDA algorithms (Fig.~\ref{fig_appx_effect_hyperparameters_ka10}, 2nd \& 4th columns).
In contrast, the VB implementation is virtually unable to infer any meaningful structure when using the default hyperparameters of Gibbs Sampling (Fig.~\ref{fig_appx_effect_hyperparameters_ka10}, 5th column).
Interestingly, when the true number of topics is unknown, we observe substantial differences in \textit{how} the two algorithms overfit the ground truth structure (Fig.~S5). 

To ensure the reliability of these findings, we repeated our analyses increasing the number of iterations 10-fold for each algorithm, obtaining identical results
(Figs.~S6, S7). 

These results confirm that the choice of default hyperparameters can bias the output of topic modeling algorithms.
More generally, however, they show how our approach can reveal intricate differences in performance which are inaccessible in standard evaluation approaches such as document classification, 
where only partial information on the inferred structure is used, e.g., the maximum in the topic-document distribution $P(t | d)$ (Fig.~S8). 

\subsection{Insights on real world corpora}
\begin{figure*}[h]
	\begin{center}
		\centerline{\includegraphics[width=1.8\columnwidth]{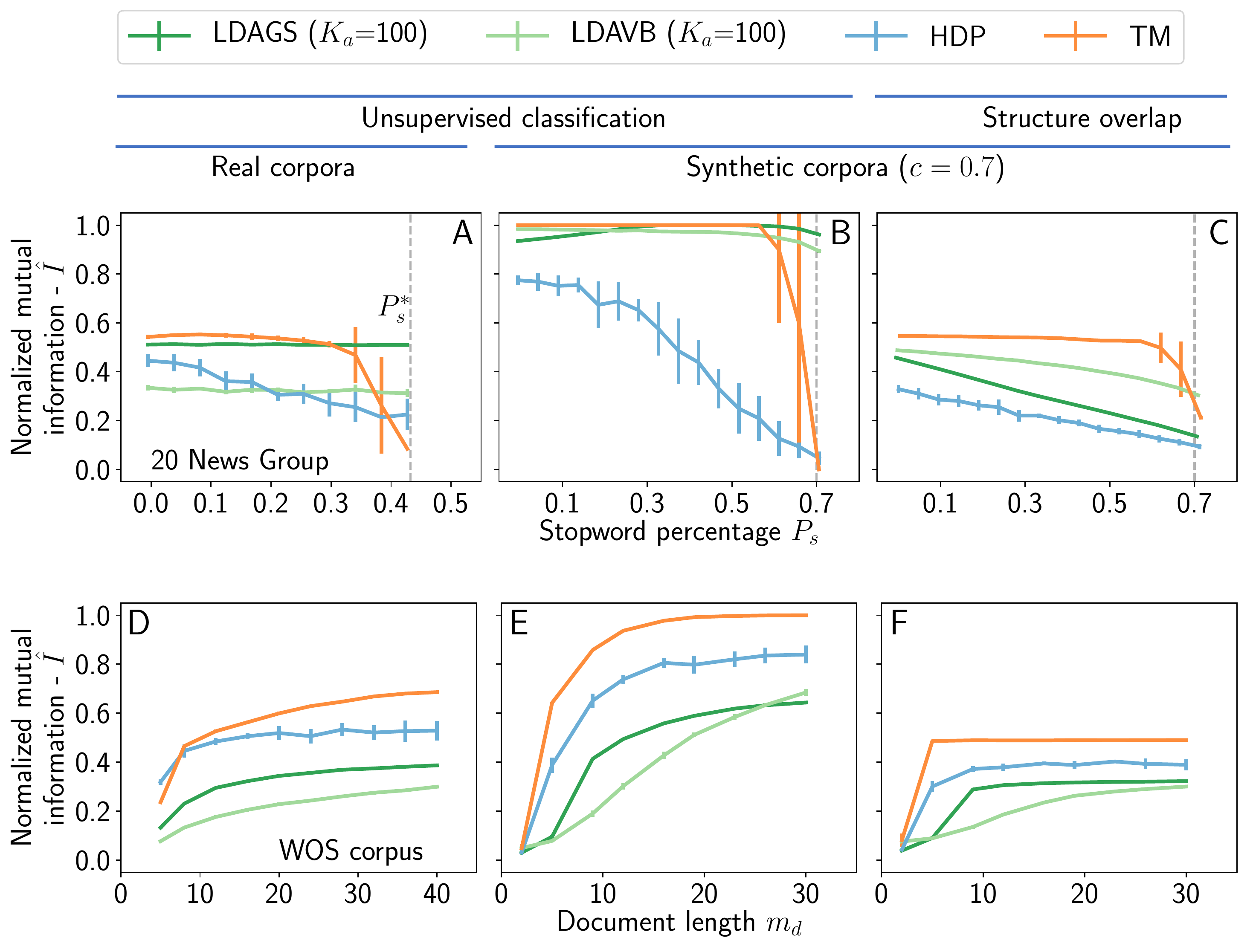}}

		\caption{
			\textbf{Performance in synthetic corpora is strongly correlated to performance in real-world corpora.}
			Comparison between 20 News Group data and synthetic corpora with $K=40$, $D=10^4$  $m_d=100$, $V=10^3$, $c=0.7$ varying the fraction of stopwords $P_s$ (top row) using $K_a=100$, and WOS data and a synthetic corpus with $K=10$, $D=10^4$, $V=10^3$, $c=0.7$ varying the document length $m_d$ (bottom row).
			(\textbf{A}, \textbf{D}) NMI from unsupervised document classification in real-world corpora.
			(\textbf{B}, \textbf{E}) NMI from unsupervised document classification in synthetic corpora.
			(\textbf{C}, \textbf{F}) NMI from structure overlap in synthetic corpora.
			While each case measures NMI (in bits), panels (A,B,D,E) compare labels of documents and panels (C,F) compare labels of word tokens.
		}
		\label{fig_linking_stopwords}
	\end{center}
	\vspace{-.1in}
\end{figure*}

%
The synthetic corpora discussed earlier constitute a simplified abstraction of the topic structure of real-world corpora.
Thus, it may not be obvious that the insights drawn from synthetic corpora will be generalizable to real-world corpora.
Therefore, we next investigate two examples supporting the hypothesis that despite its simplicity the synthetic corpus not only allows to make predictions on the performance of topic modeling algorithms in similar real-world corpora, but it also provides additional insights as to why different algorithms perform differently on distinct corpora~(Fig.~\ref{fig_linking_stopwords}).

We measure performance in real-world corpora in an unsupervised classification task using human-assigned document labels as a ground truth proxy. In analogy to the approach in  Eqs.~(\ref{eq.ptt},\ref{eq.nmi}) we quantify the correspondence between external and inferred document labels using the normalized mutual information (see \textit{Supplementary Material} Sec.~S4). 

\paragraph{Stopwords.}

While many practitioners remove stopwords from corpora prior to analysis, there is no consensus on the effect of stopwords on the performance of topic modeling algorithms~\cite{Zaman2011,Schofield2017}. 
We thus investigate the effect of stopwords using the 20 News Group (20NG) dataset motivated by the fact that it exhibited the strongest dependence of performance on stopword shown in Fig.~\ref{fig_compare_real}.
Using the English stopword list from MALLET~\cite{mccallum2002mallet}, we estimate that about 43\% of word tokens in the 20NG corpus are stopwords. 
For our analysis, we remove at random a given fraction of these tokens.
We find that performance of topic modeling algorithms varies but generally increases as we decrease the fraction of stopwords (Fig.~\ref{fig_linking_stopwords}A).

We construct a synthetic corpus with  $c=0.7$ (we obtain similar results with different values, Fig.~S9), 
$K=40$, and a varying fraction $P_s$ of stopwords.   
Measuring performance by unsupervised document classification, we find the same pattern as for the real corpus (Fig.~\ref{fig_linking_stopwords}B).
In contrast, measuring performance as the overlap between planted and inferred structure yields substantial differences, which reflect the additional detail provided by the structure overlap  (Fig.~\ref{fig_linking_stopwords}C).
Considering the inferred topic distributions (Fig.~S10), 
we find that LDAVB infers a pure topic-document distribution, 
assigning most of the uncertainty to the word-topic distribution and correctly identifying most of the stopwords, 
while LDAGS assigns most of the uncertainty to the topic-document distribution,
trying to infer a pure word-topic distribution resulting in overfitting the stopwords and assigning them to inferred topics.
In document classification, most of this information remains invisible, leading to indistinguishable results for the two algorithms.

\paragraph{Document length.}
It has been reported that topic models have low performance on corpora of short document, such as Twitter posts~\cite{Hong2010}. However, the effect of document length on the performance of topic models is still not well characterized~\cite{Tang2014}.
We thus investigate the effect of text length by considering only the first $m_d$ words of each document in the Web of Science (WOS) dataset, a collection of 40,526 scientific articles (title and abstract) from 7 academic areas.
Prior to analysis, we removed all stopwords (using the stopword list from MALLET~\cite{mccallum2002mallet}).
We find that performance improves with increasing document length~(Fig.~\ref{fig_linking_stopwords}D); yet, the ranking of the models' performance remains virtually unchanged.

We construct a synthetic corpus with similar properties fixing $c=0.7$ (we obtain similar results with different values, Fig.~S11) 
and $K=10$ and varying the length $m_d$ of each document.   
For both measures of performance, classification (Fig.~\ref{fig_linking_stopwords}E) and structure overlap (Fig.~\ref{fig_linking_stopwords}F) we qualitatively reproduce the findings on the real corpus.
In particular, we recover the same ranking for the performance of topic models.

\section{Discussion}

Our study illustrates how the use of synthetic corpora can lead to new insights on topic model performance unattainable when only studying real-world corpora. 
Our approach allows us to systematically investigate the effect of both individual properties of the corpus (document length, stopwords, etc.) and parameters of the topic modeling algorithms (assumed number of topics, hyperparameters, etc.).
For example, our analysis reveals that
({\it i}) the number of topics determined by popular non-parametric approaches (such as HDP) cannot be relied upon; 
({\it ii}) there exist fundamental limits to algorithms' ability to infer a topic structure.
and ({\it iii}) the default hyperparameter settings induce a substantial bias in the inferred solutions of different implementations of the same topic model.
Most importantly, we demonstrate the practical relevance of our approach by showing that relative performance in synthetic corpora predicts relative performance in real-world corpora.

While these results raise more questions than they can answer, we believe that our proposed framework offers a complimentary approach to gain a better understanding of topic modeling algorithms.
In particular, it allows us to systematically identify strengths and weaknesses of topic modeling algorithms in different applications and under different conditions allowing for more informed choices among a large number of available algorithms.

Unarguably, the presented synthetic corpora are far from the complexity of real-world corpora.
However, our framework provides enough flexibility to accommodate different 
features such as burstiness, syntax, or structures beyond the bag-of-words model (phrases, sentences, etc.) in
future studies with increasing complexity of the synthetic corpora. 

\subsubsection*{Acknowledgements}
L.A.N.A. thanks the John and Leslie McQuown Gift and support from Department of Defense Army Research Office (Grant Number W911NF-14-1-0259).

\clearpage


\bibliography{aistats2019_topic_model,aistats2019_topic_model_package}

\end{document}


%

%

\twocolumn[

\aistatstitle{ Supplementary Material for the manuscript:\\ ``A new evaluation framework for topic modeling algorithms based on synthetic corpora''}

\aistatsauthor{ Hanyu Shi $^{1}$ \And Martin Gerlach$^{1}$ \And  Isabel Diersen$^1$ \And Doug Downey$^2$ \And Lu\'{i}s A. N. Amaral$^{1, *}$   }

\aistatsaddress{  $^1$Department of Chemical and Biological Engineering, $^2$Department of Electrical Engineering and Computer Science \\
Northwestern University \\ 
Evanston, Illinois 60208, USA}
]

\renewcommand{\theequation}{S\arabic{equation}}
\renewcommand{\thefigure}{S\arabic{figure}}
\renewcommand{\thetable}{S\arabic{table}}
\renewcommand{\thesection}{S\arabic{section}}

\setcounter{equation}{0}  
\setcounter{figure}{0}
\setcounter{table}{0}
\setcounter{section}{0}

 \renewcommand{\refname}{Supplementary References}
    \makeatletter
    \renewcommand\@biblabel[1]{[S#1]}
    \renewcommand\citeform[1]{S#1}

\section{Real world corpora}
\label{appx_real_corpora}
We use $4$ datasets and apply different filtering strategies yielding 8 different real world corpora each with $D$ documents, $N$ tokens, and $C$ category labels, as shown in Table~\ref{table_sm01_real_corpora}.

\begin{table*}[h]
\begin{centering}
	\caption{\textbf{Details for real-world corpora.}}
	\vspace{.1in}
	\begin{tabularx}{\textwidth}{XlXX}
		\hline 
		\hline 
		Dataset &  Variation & Filtering & Characteristics \tabularnewline
		\hline 
		Reuters-21578~\cite{reuters21578} & 1 & Only documents from the 10 largest categories & $D$=7,518; $N$=775,771; $C$=10  \tabularnewline
		& 2 & Only documents from categories with more than 10 documents & $D$=8,559; $N$=936,004; $C$=41  \tabularnewline
		\hline
		RCV1~\cite{rcv1} & 1 & Only documents with one category label; subsample ~10\% of documents from the largest category  & $D$=3,070; $N$=574,249; $C$=4  \tabularnewline
		& 2 & Only documents with two category labels; subsample 10\% of documents & $D$=20,474; $N$=2,917,939; $C$=54  \tabularnewline
		\hline
		Web of Science~\cite{tm2017} & 1 & None & $D$=40,526; $N$=3,828,735; $C$=7  \tabularnewline
		& 2 & Only keep the first 20 tokens of each document  & $D$=40,526; $N$=808,672; $C$=7  \tabularnewline
		\hline
		20 News Group~\cite{20ng} & 1 & Remove all words with less than 3 characters & $D$=18,803; $N$=3,831,559; $C$=20  \tabularnewline
		& 2 & Same as in variation 1 and remove all stopwords from list given in Ref.~\cite{mccallum2002mallet}   & $D$=18,799; $N$=2,654,710, $C$=20;   \tabularnewline
		\hline 
		\hline 
	\end{tabularx}
\end{centering}
\label{table_sm01_real_corpora}
\end{table*}

\section{Topic model algorithms}
\label{appx_topic_models}

We use the following topic modeling algorithms (shown in Table~\ref{table_sm02_topic_models}) with default parameter settings unless stated otherwise.

\begin{table*}[h]	
\begin{centering}
	\caption{\textbf{Details for topic modeling algorithms.}}
	\vspace{.1in}
	\begin{tabularx}{\textwidth}{XXX}
		 \hline 
		 \hline
		Topic Model &  Implementation & Default hyperparameter values \tabularnewline
		\hline 
		Gibbs Sampling LDA (LDAGS)  \cite{Griffiths2004} & Mallet \cite{mccallum2002mallet} & $\alpha=5/K_a$, $\beta=0.01$  \tabularnewline
		\hline 
		Variational Bayes LDA (LDAVB) \cite{Blei2003} & gensim \cite{Rehurek10} &$\alpha=1/K_a$, $\beta=1/K_a$ \tabularnewline
		\hline 
		Hierarchical Dirichlet Processes (HDP) \cite{Teh2006} & From work of \cite{hdp2017} & n.a. \tabularnewline
		\hline 
		TopicMapping (TM) \\ \cite{Lancichinetti2015} & From work of \cite{tm2017} & n.a. \tabularnewline
		\hline 
		\hline
	\end{tabularx}
\end{centering}
\label{table_sm02_topic_models}
\end{table*}

LDA requires hyperparameter values for the topic-document distribution, a $K_a$-dimensional vector $\vec{\alpha}=(\alpha_j)_{j=1,\ldots,K_a}$ (where $K_a$ is the assumed number of topics), and the word-topic distribution, a $V$-dimensional vector $\vec{\beta}=(\beta_w)_{w=1,\ldots,V}$ (where $V$ is the size of the vocabulary).
We assume symmetric priors, i.e. $\alpha_j=\alpha$ and $\beta_w=\beta$, such that the hyperparameters are fully determined by the scalar parameters $\alpha$ and $\beta$.
For LDAVB we use the default values of the gensim implementations. 
For LDAGS we use the default values of the gensim-wrapper of the mallet implementation.

\section{Usage of synthetic corpora in previous studies}

Different types of synthetic corpora in previous studies are given in Table~\ref{table_sm03_previous_synthetic}. 
As we can see, in previous research a large portion of synthetic corpora are generated directly from LDA.

\begin{table*}[h]	
\begin{centering}
\caption{\textbf{Usage of synthetic corpora in previous studies.}}
\vspace{.1in}
	\begin{tabularx}{\textwidth}{XXX}
	\hline 
	\hline
	Reference &  Synthetic Corpora & Corresponding evaluation metric \tabularnewline

	\hline 
	\cite{Mukherjee2009} &  Generated from LDA & Likelihood; Variational free energy \tabularnewline
	\hline 
	 \cite{Newman_2009} & Generated from LDA & L1-norm between true and inferred word-topic distribution \tabularnewline
	\hline 
	\cite{Wallach2009a} &  Generated from LDA & Held-out likelihood \tabularnewline
	\hline 
	\cite{Mimno-Blei2011} &  Generated from LDA & Check hypothesis that words and documents are independent given the topic using mutual information \tabularnewline
	\hline 
	\cite{Taddy2012} & Generated from LDA &  Compare entries (and residuals) in $\theta_k$ (defined as the distribution over words for each topic) between true topics and inferred topics \tabularnewline
	\hline 
	\cite{Tang2014} & Generated from LDA & Posterior contraction analysis of the topic polytope \tabularnewline
	\hline 
	\cite{Hsu_NIPS2016} & Generated from LDA & Use the synthetic corpora to test inferred number of topics \tabularnewline
	\hline 
	\cite{Minka2002} &  Multinomial with 5 equiprobable words & Likelihood; Classification \tabularnewline
	\hline 
	\cite{Griffiths2004} &  “Bar”-data (5x5 grid)  & Visual comparison \tabularnewline
	\hline 
	\cite{Andrzejewski2009} &  Small size synthetic corpora based on their proposed topic model (LDA with Dirichlet Forest Priors)  & Visual inspection of the word-document matrix \tabularnewline
	\hline 
	\cite{AlSumait2009} &  Small size synthetic corpora: 6 samples of 16 documents from three static equally weighted topic distributions. On average, the document size was 16 words.  & Topic signficance score (similar to topic coherence) \tabularnewline
	\hline 
	\cite{Arora2013, Arora2016} & Semi-synthetic data (train parameters of a model on a real corpus, then use the model to generate synthetic data) & Training time; L1-error between true and inferred matrix A (defined as the word-topic matrix). \tabularnewline
	\hline 
	\cite{Lancichinetti2015} & “Language” data with non-overlapping topics  & L1-norm between true and inferred $p(d|t)$ \tabularnewline
	\hline 
	This work & A flexible framework that could include a range of topic structure and realistic features  & Measure the overlap between the planted and the inferred topic labels on the token level \tabularnewline
	\hline 
	\hline
	\end{tabularx}
\end{centering}
\label{table_sm03_previous_synthetic}
\end{table*}

\section{Document classification}
\label{appx_classification}

In practical applications, topic models are often used to find documents of similar topical content in an unsupervised fashion.
In this spirit we quantify the performance of topic models by checking how much the inferred topic distributions reflect the assignment of human-labeled categories in real world corpora.

More specifically, we fit a topic model to the entire corpus and obtain the inferred topic-distribution of each document, $P(t | d)$.
Identifying each topic with a category, we predict the category-membership of each document, $s_d$, from the topic with maximum probability~\cite{Xie2013}
\begin{equation}
s_d = \underset{t}{\arg\max} P(t | d)
\end{equation}
Comparing this with the given metadata-category, $r_d$, we can construct a confusion matrix 
\begin{equation}
p_{s,r} \equiv \frac{1}{D}\sum_d \delta_{s,s_d} \cdot \delta_{r,r_d},
\end{equation}
which yields the fraction of documents that have metadata-category $r$ and predicted category $s$.
With confusion matrix  $p_{s,r}$, we can quantify the performance of the topic model in the classification task using the normalized mutual information.

\section{Generation of synthetic benchmark corpora}
\label{appx_generate_synthetic}

With the distributions $P(w|t)$ and $P(t|d)$ described in the main text Sec.~3.1, 
we can generate the synthetic benchmark corpora from distributions $P(w|t)$ and $P(t|d)$ according to the generative process.
One major advantage of our work is that our approach allows for the inclusion of many realistic features, such as Zipfian distribution, stopword, and burstiness, as described below.

\paragraph{Zipfian word-frequency distribution.}
One of the most well-known statistical laws in language is the so-called Zipf's law~\cite{zipf1936psycho}, which states that the frequency $f$ of the $r$-th most frequent word is given by a power-law with exponent $\gamma > 1$:
\begin{equation}
f(r) \propto r^{-\gamma}
\end{equation}
We incorporate a Zipfian distribution by accommodating any global word-frequency distribution $P(w)$ as an average over all topics, i.e. $P(w) = \sum_t P(w,t)=\sum_t P(w | t)P(t)$ where $P(t)$ is the size of topic $t$.

\paragraph{Stopwords.}
An important statistical property of real texts is the generic appearance of stopwords. 
While there is no general agreed-upon definition, in the context of topic modeling this usually refers to very common words (such as ``the,'' ``and,'' etc.) which are considered not informative in inferring topical structure related to semantics. 
In practice, these words are typically removed from a corpus using a pre-specified list of stopwords, however, there exist differing opinions on the effect of stopwords in the result of the quality of inferred topic models~\cite{Zaman2011,Schofield2017}. 

We model stopwords as words which have the same probability of appearance in any topic, i.e. $P(w | t) = P(w)$.
Varying the fraction of stopwords (of unique words in the vocabulary) by a parameter $P_s \in [0,1]$ allows us to investigate the robustness of a topic model with respect to these non-informative words.

\paragraph{Burstiness.}
The phenomenon of burstiness refers to non-stationarity in the usage of words~\cite{KATZ1996, Altmann2009}, that is a word is more likely to occur in a text after its first occurrence.

We incorporate burstiness following approaches proposed in Refs.~\cite{Madsen2005,Doyle2009} using Dirichlet-distributions.
Given a topic $t$, instead of drawing from a fixed word-topic distribution $P(w | t)$, we obtain a different word-topic distribution in each document which is drawn from a $V$-dimensional Dirichlet distribution with concentration parameter $a_c$, i.e.  $P_d(w | t) \sim Dir_V( a_c \cdot P(w | t) )$.
This means that the smaller $a_c$ the more ``bursty'' the synthetic corpora. 
For example, in the limiting case $a_c \rightarrow 0$ ($a_c \rightarrow \infty$) the word-topic distribution in each document will contain only one word with non-zero probability (the original global word-topic distribution from the non-bursty case).

\section{Supplementary figures}
\label{sec.suppl.figures}

\begin{figure*}[h]
	\begin{center}
		\includegraphics[width=1.8\columnwidth]{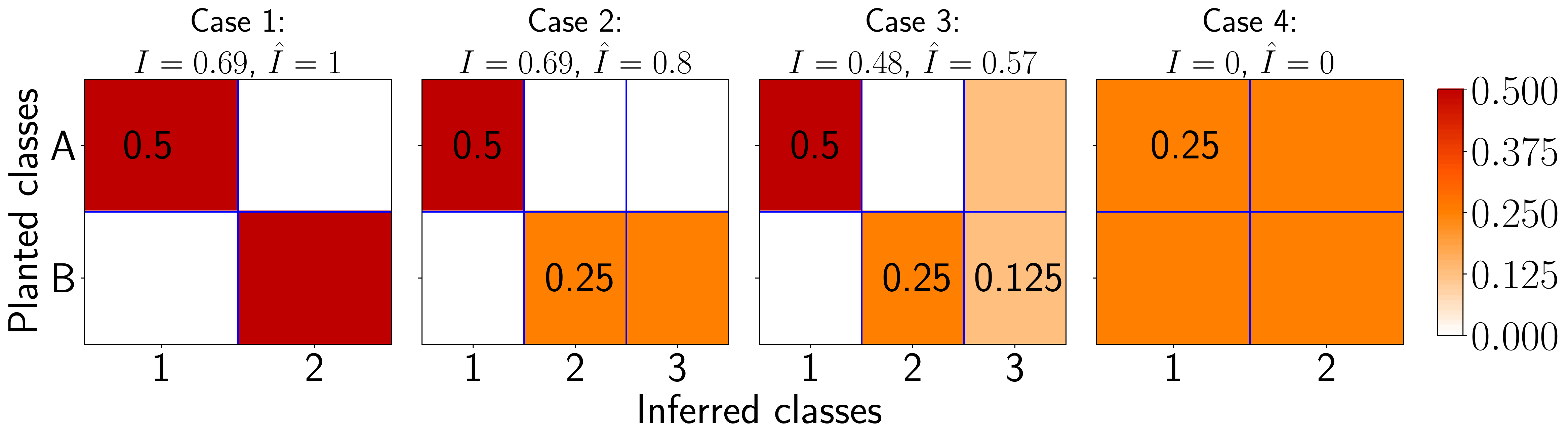}
		\vskip -0.2in
		\caption{
			\textbf{Quantifying overlap using the normalized mutual information.}
			Unnormalized, $I$, and normalized mutual information, $\hat{I}$,  for different examples of confusion matrices $p_{t,t'}$ (cases 1-4).
			Number indicate the values of the confusion matrix according to color.
		}
		\label{fig_example_NMI_MI}
	\end{center}
\end{figure*}

As an example in Fig.~\ref{fig_example_NMI_MI}, consider two planted classes in the synthetic benchmark,  where 50\% of the tokens belong to each planted class, we obtain different values $\hat{I}$ depending on the inferred structure:
({\it 1}) If all tokens are correctly assigned into two inferred classes yielding a perfectly diagonal confusion matrix $p_{t,t'}$, this leads to $I=log(2)$ and $\hat I = 1$;
({\it 2}) In case one of the inferred classes gets split into two equal-sized classes, we get the same $I$, but a smaller value for $\hat I$;
({\it 3}) If one of the smaller inferred classes is uninformative with respect to the planted classes, this leads to a further reduction of $I$ and $\hat I$;
({\it 4}) If the tokens are just randomly assigned to two inferred classes, this yields  $I=0$ and $\hat I = 0$.

\begin{figure*}[h]
	\begin{center}
		\includegraphics[width=1.8\columnwidth]{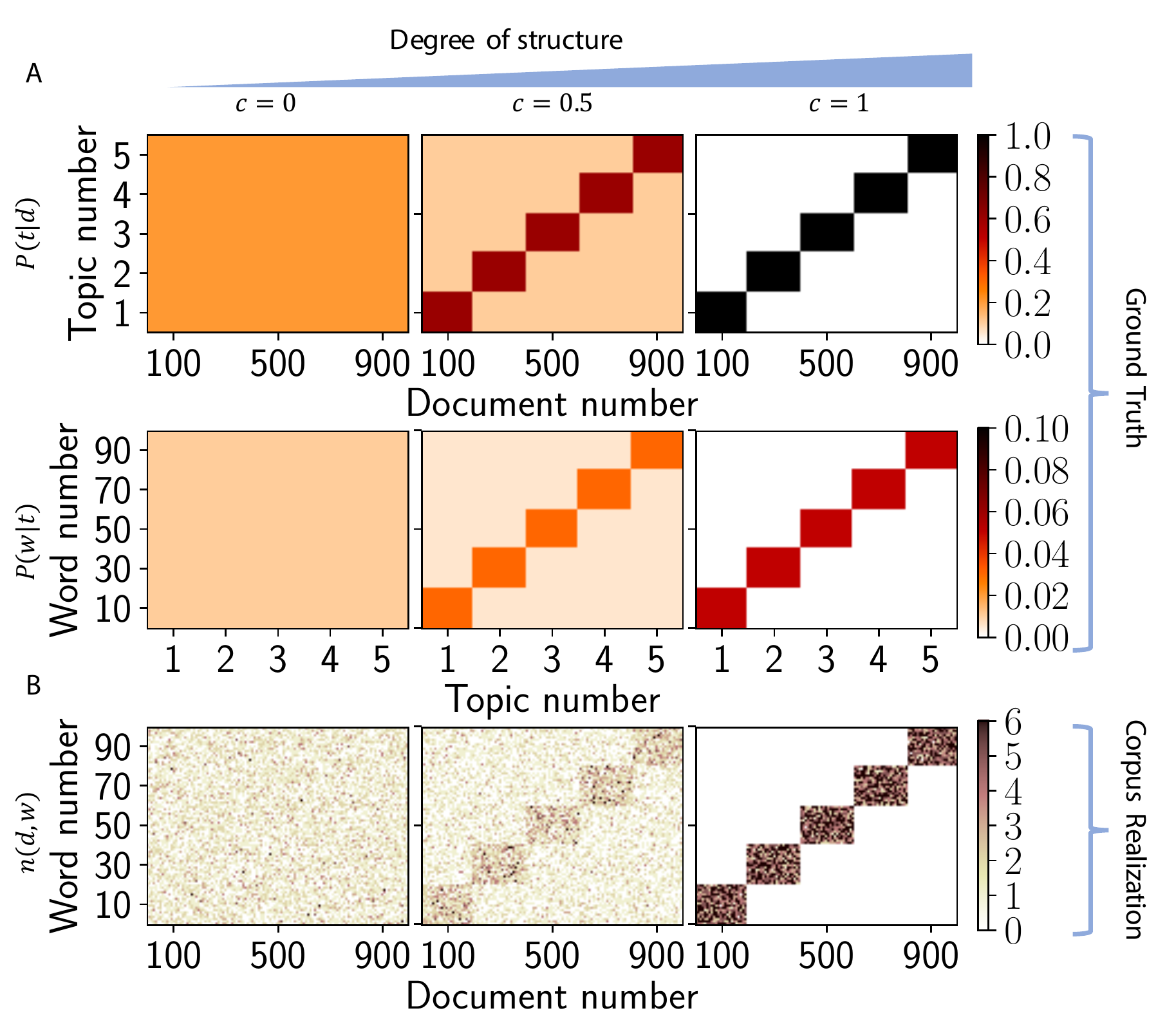}
		\vskip -0.1in
		\caption{\textbf{ Synthetic benchmarks with known ground truth from the generative process of topic models.} 
			Three synthetic benchmark corpora with equal size but different degrees of structure $c \in \{ 0,0.5,1 \}$ (left, middle, right column).
			\textbf{(A)} Ground truth topic distributions $P(t | d)$ and $P(w | t)$.
			\textbf{(B)} Resulting observable corpus showing the number of times word $w$ appears in document $d$, $n(d,w)$.
			For all panels, the number of topics is $K=5$; there are $V=100$ words in the vocabulary; and each corpus contains $D=1,000$ document with length of $m_d=100$.
		}
		\label{fig_synthetic_copora}
	\end{center}
\end{figure*}

In Fig.~\ref{fig_synthetic_copora} we show the resulting synthetic corpora for the random ($c=0$), mixed ($c=0.5$), and ordered ($c=1$) case.
While for $c=0$ the topics are not distinguishable in the ground truth topic distributions, $c>0$ yields a  block-diagonal structure (Fig.~\ref{fig_synthetic_copora}A).
Looking at the  empirically observed corpus in the form of the counts $n(d,w)$, i.e., the number of times word $w$ appears in document $d$, 
words are distributed randomly across all documents for $c=0$, while increasing $c$ leads to a higher concentration of words in certain documents reflecting the increasing degree of structure (Fig.~\ref{fig_synthetic_copora}B).

\begin{figure*}[h]
	\begin{center}
		\centerline{\includegraphics[width=1.8\columnwidth]{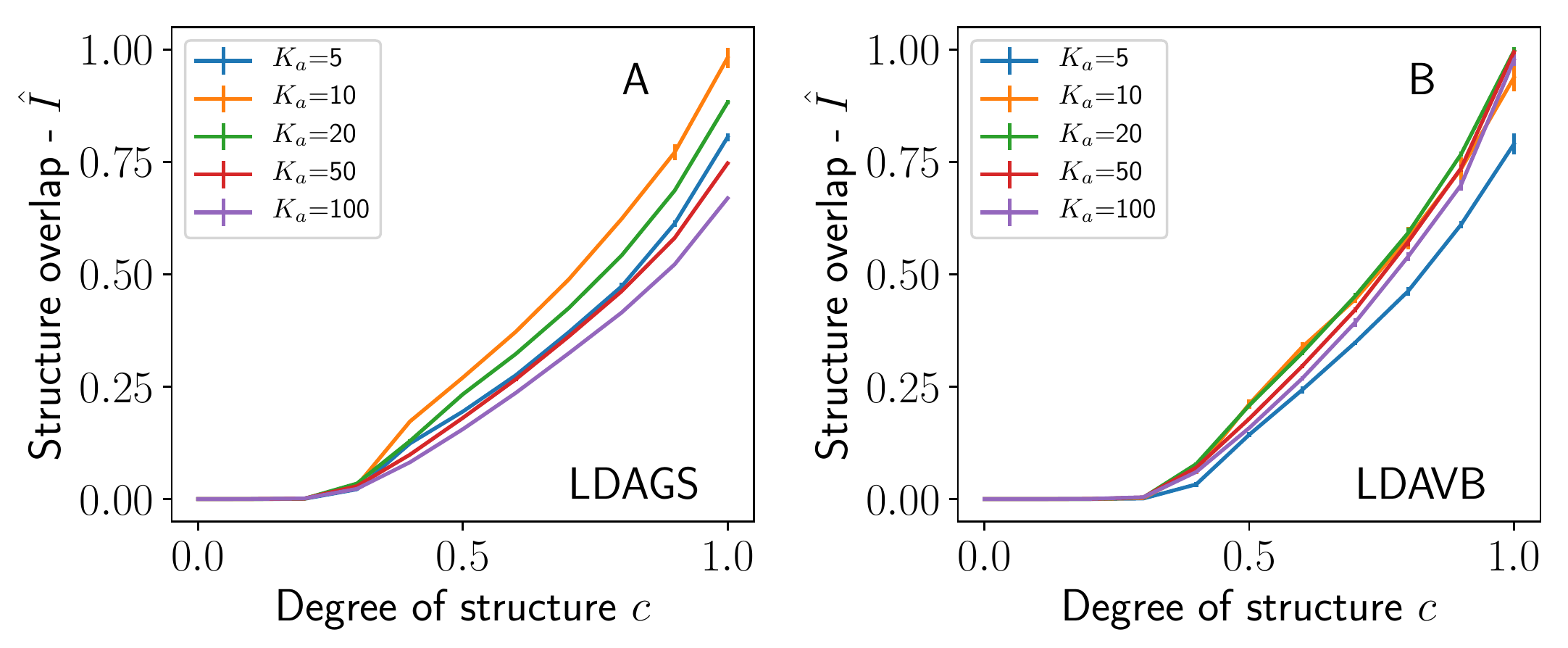}}
		\caption{ 
			\textbf{Both overfitting and underfitting reduce the performance of LDA models.} 
			Normalized mutual information, $\hat I$, between planted and inferred structure as a function of the structure parameter $c$ varying the assumed number of topics $K_a \in \{ 5,10,20,50,100\}$.
			\textbf{(A)} Gibbs Sampling LDA.
			\textbf{(B)} Variational Bayes LDA.
			For the synthetic benchmark corpora, we set the planted number of topics as $K=10$, the vocabulary as $V=1,000$, and the number of documents as $D=10,000$ each with length $m_d= 100$.
			In the experiment we use different assumed numbers of topics (5, 10, 20, 50, 100) for LDA methods.
			The curves denote averages (and $\pm$ one standard deviation) over $10$ realizations.
		}
		\label{fig_appx_change_lda_kset}
	\end{center}
\end{figure*}

For algorithms such as LDA one needs to specify the number of topics for fitting the topic model.
While in the synthetic corpus we know the true number of topics, in practice, this value is unknown.
Therefore, we investigate the effect of over- and under-fitting by varying the assumed number of topics, $K_a$, for LDA in a synthetic corpus with 10 planted topics (Fig.~\ref{fig_appx_change_lda_kset}).

\begin{figure*}[h]
	\begin{center}
		\centerline{\includegraphics[width=1.8\columnwidth]{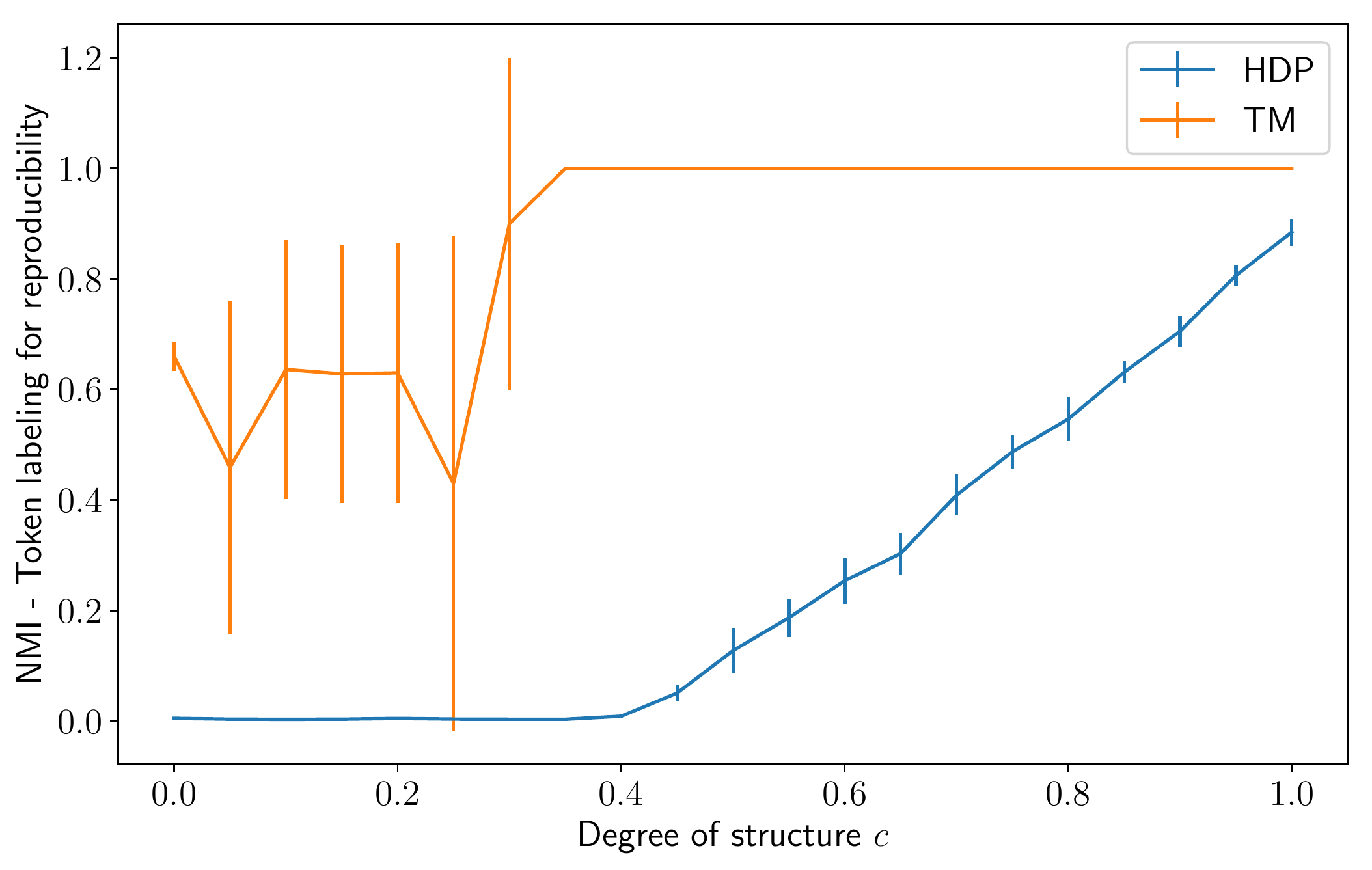}}
		\caption{ 
			\textbf{Compare the reproducibility of two nonparametric topic modeling algorithms, HDP and TM, based on token labeling comparison.}
			Synthetic corpora were generated with $K=10$ topics, $D=10^4$ documents, document length $m=100$, and vocabulary size $V=10^3$.
			The lines (error bars) denote averages (one standard deviation) estimated from $10$ realizations.
		}
		\label{fig_appx_reproducibility_hdp_tm}
	\end{center}
\end{figure*}

Fig.~\ref{fig_appx_reproducibility_hdp_tm} compares the reproducibility of HDP and TM. 
There are 10 repetitions for each data points. 
In each repetition, only one synthetic benchmark corpus is generated. 
A topic model will be run on this corpus twice. 
The inferred token topics form the two experiments of the topic model will be compared.

\begin{figure*}[h]
	\vskip -0.1in
	\begin{center}
		\centerline{\includegraphics[width=1.8\columnwidth]{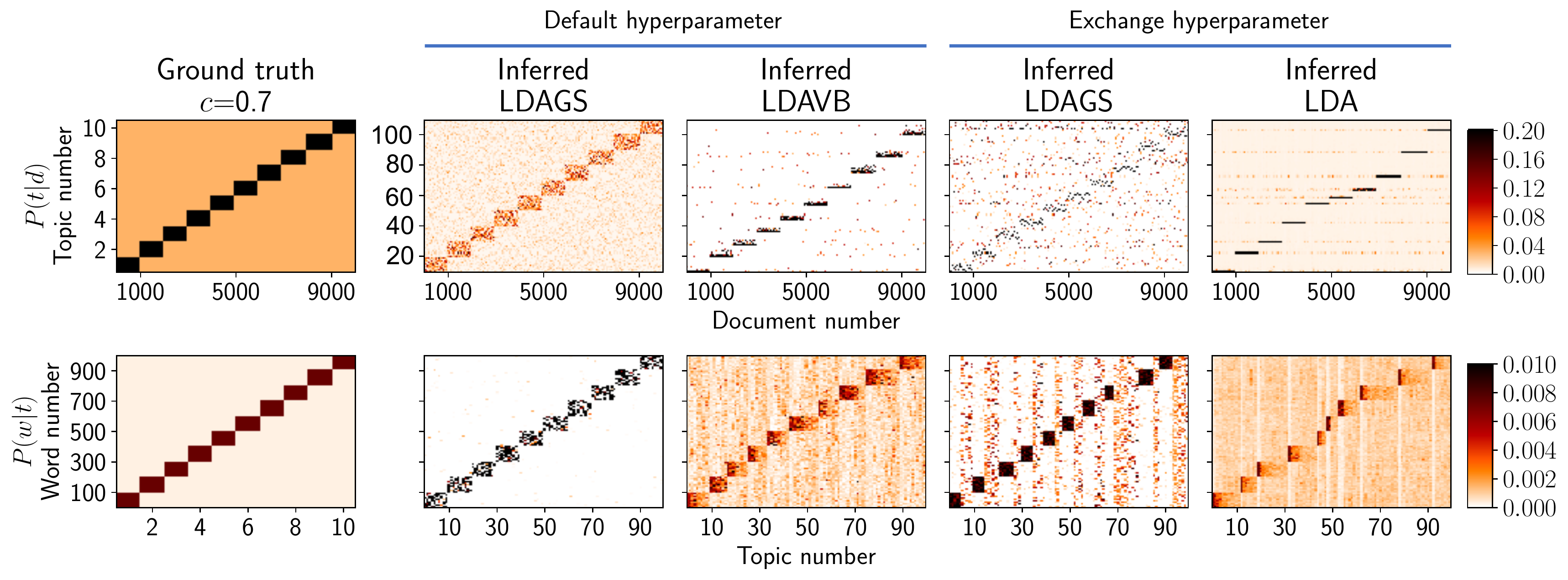}}
		\caption{ 
			\textbf{Hyperparameters bias the inferred topic structure of different algorithms of LDA models.} 
			Comparison of topic distributions $P(t|d)$ (top row) and $P(w|t)$ (bottom row) from the planted and inferred structure from LDAGS and LDAVB using two different sets of hyperparameters: original defaults as defined in each implementation (middle panels) and defaults from the other implementation, respectively (right panels).
			Same parameters as in Fig.~3 
			fixing $c=0.7$ and using $K_a=100$.
		}
		\label{fig_appx_effect_hyperparameters_ka100}
	\end{center}
	\vskip -0.3in
\end{figure*}

In Fig.~\ref{fig_appx_effect_hyperparameters_ka100}, we show the planted and inferred $P(t | d)$ and $P(w | t)$ by the implementation of different algorithms for LDA, using $K_a=100$ as the assumed number of topics.

\begin{figure*}[h]
	\begin{center}
		\centerline{\includegraphics[width=1.8\columnwidth]{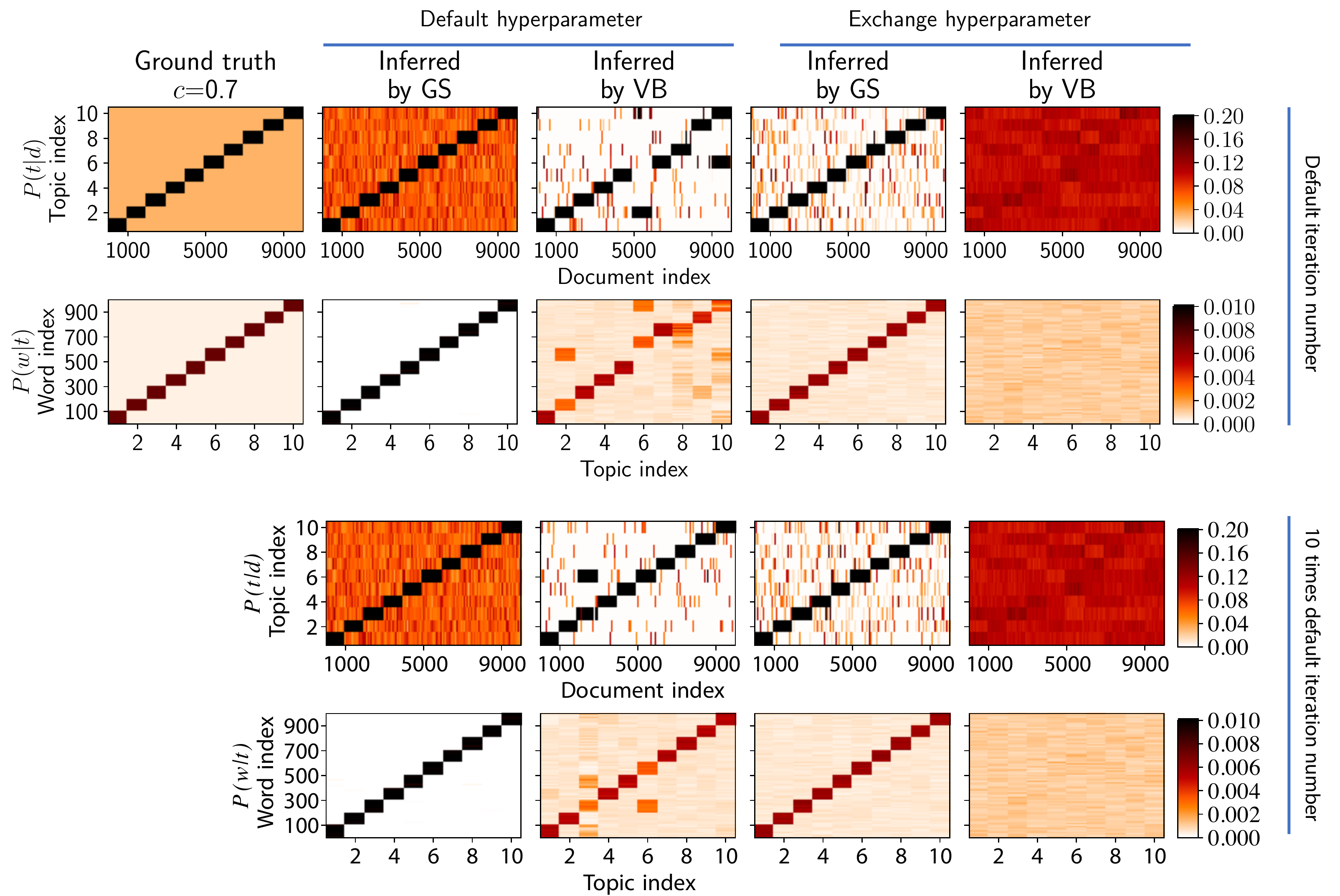}}

		\caption{ 
			\textbf{Topic models converge with the default iteration setting for $K_a=10$.}  
			Inferred topic distributions $P(t|d)$ and $P(w|t)$ as in Fig.~4 
			for Gibbs Sampling LDA and Variational Bayes LDA with different hyperparameter settings comparing the case where we use the default number of iterations (top two columns) with the case where we increase the number of iterations 10-fold (bottom two columns).
		}
		\label{fig_appx_effect_hyperparameters_DiffIterSet_ka10}
	\end{center}
\end{figure*}

\begin{figure*}[h]
	\begin{center}
		\centerline{\includegraphics[width=1.8\columnwidth]{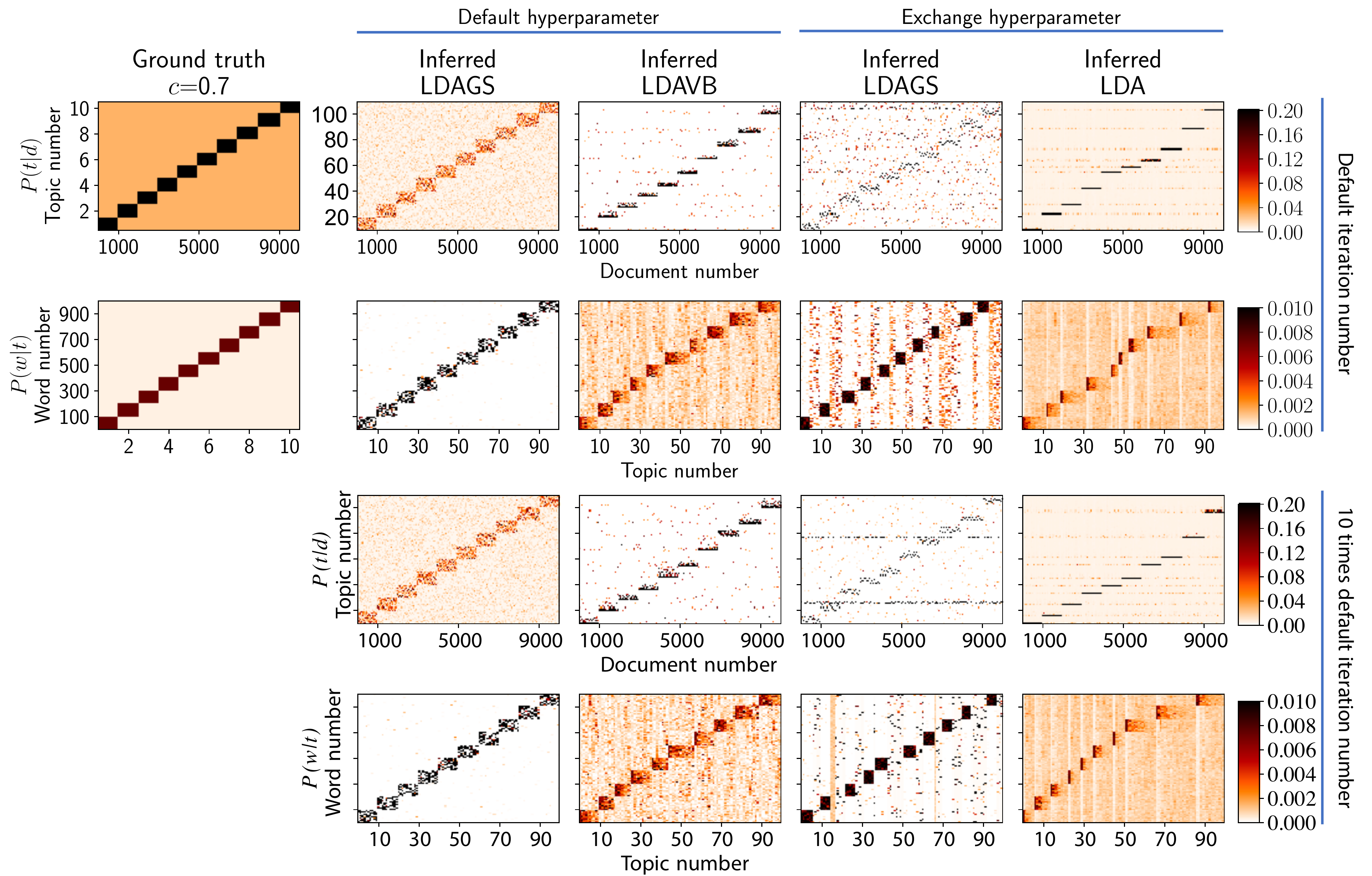}}
		\caption{ 
			\textbf{Topic models converge with the default iteration setting for $K_a=100$.}  
			Inferred topic distributions $P(t|d)$ and $P(w|t)$ as in Fig.~4 
			for Gibbs Sampling LDA and Variational Bayes LDA with different hyperparameter settings comparing the case where we use the default number of iterations (top two columns) with the case where we increase the number of iterations 10-fold (bottom two columns).
		}
		\label{fig_appx_effect_hyperparameters_DiffIterSet_ka100}
	\end{center}
\end{figure*}

In Fig.~\ref{fig_appx_effect_hyperparameters_DiffIterSet_ka10} and Fig.~\ref{fig_appx_effect_hyperparameters_DiffIterSet_ka100}, we confirm that the solutions for LDA obtained in Fig.~4 
and Fig.~\ref{fig_appx_effect_hyperparameters_ka100} have converged with respect to the number of iterations in each topic model.

\begin{figure*}[h]
	\begin{center}
		\centerline{\includegraphics[width=1.8\columnwidth]{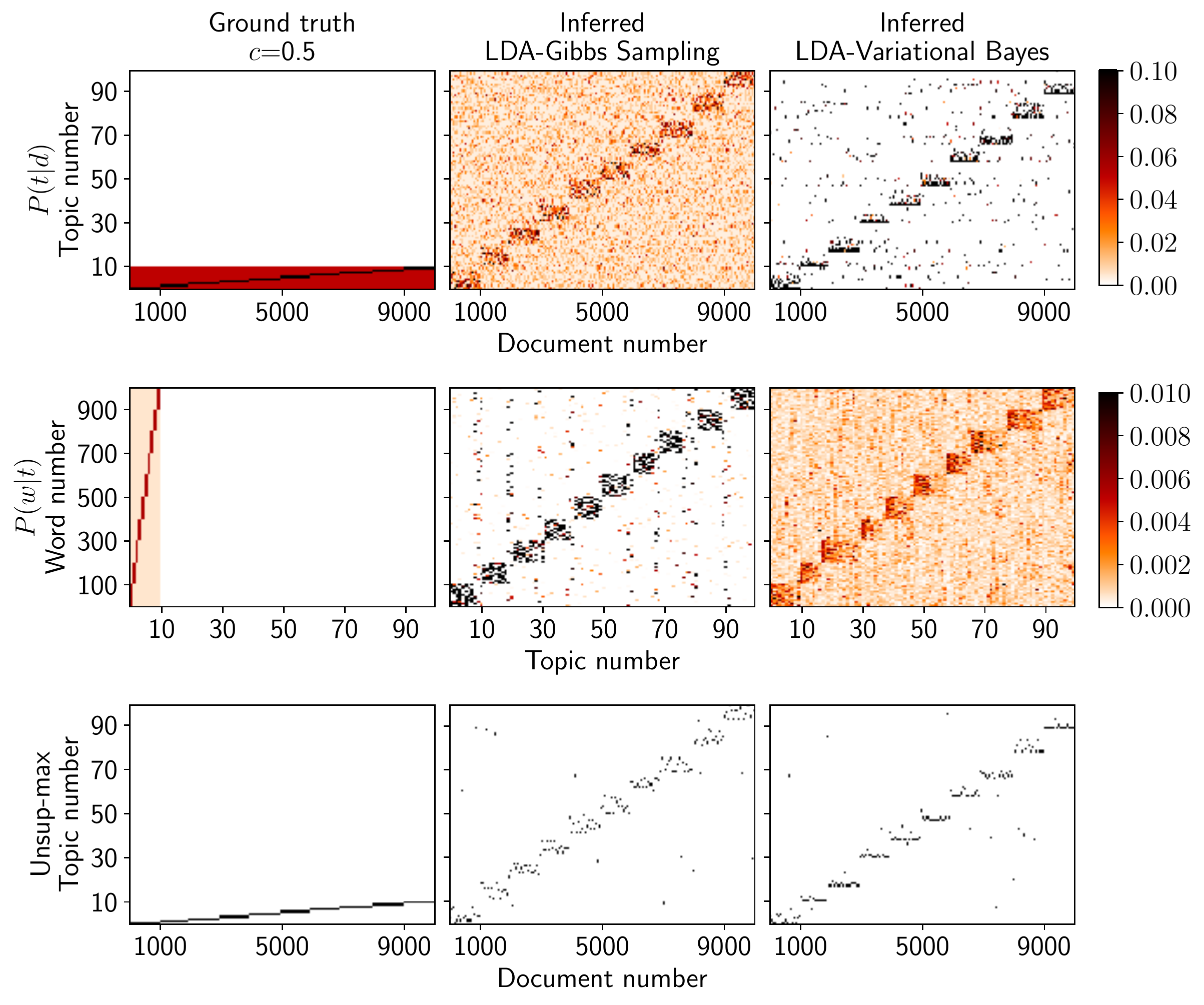}}
		\caption{ 
			\textbf{Document classification overlooks information of the inferred topic structure.}
			Comparison of the topic-document distribution  $P(t|d)$ (top row), the word-topic distribution $P(w|t)$ (middle row), and the predicted topic in unsupervised document classification $\underset{t}{\arg \max}P(t|d)$ (bottom row) for three cases:
			Ground truth as planted in the synthetic corpus (left column), inferred from Gibbs Sampling LDA (middle column), and inferred from Variational Bayes LDA (right column).
			For the synthetic benchmark corpora, we set parameters as $K=10$ topics, $D=10,000$ documents each of length $m_d=100$, $V=10^3$ as vocabulary size, and $c=0.5$ for the degree of structure. 
			For LDA models, we use $K_a=100$ as the assumed number of topics.
		}
		\label{fig_appx_compare_LDA_max}
	\end{center}
\end{figure*}

In Fig.~\ref{fig_appx_compare_LDA_max} we compare the planted and inferred topic distributions $P(t|d)$ and $P(w|t)$ and show how they provide a more detailed view on the performance of a topic model than obtained from document classification tasks.

\begin{figure*}[h]
	\begin{center}
		\centerline{\includegraphics[width=1.6 \columnwidth]{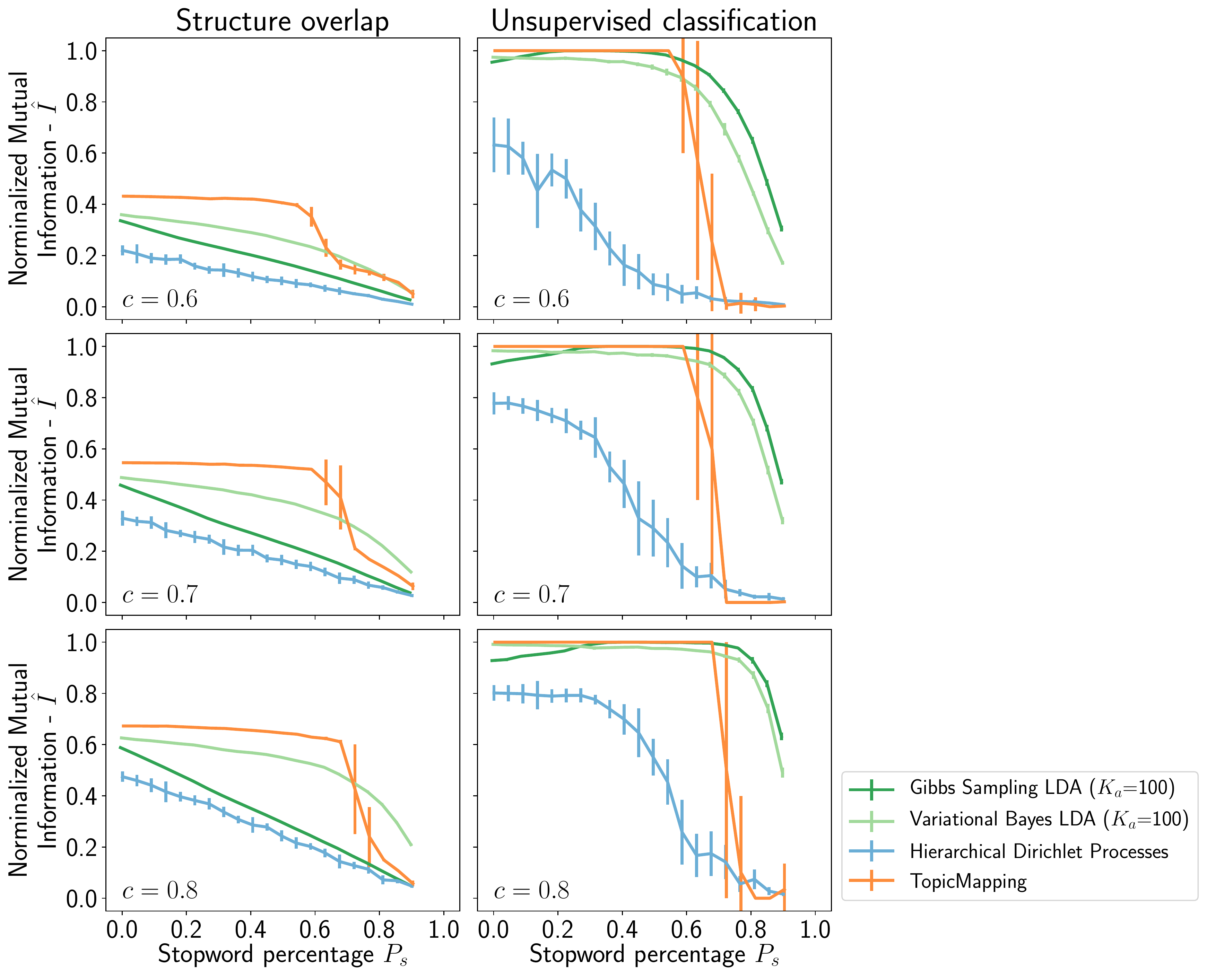}}
		\caption{
			\textbf{Varying the degree of structure does not effect results on stopword dependency in synthetic corpora.}
			Normalized mutual information, $\hat{I}$, as measured by structure overlap (left column) and unsupervised document classification (right column) as in Fig.~5 
			(E, F) varying the degree of structure:
			$c=0.6$ (top row), $c=0.7$ (middle row), and $c=0.8$ (bottom row).
		}
		\label{fig_changeC_changestopwords_compare}
	\end{center}
\end{figure*}

In Fig.~\ref{fig_changeC_changestopwords_compare} we show that the results from Fig.~5 
(B, C)  investigating the effect of stopwords on the performance of topic models in synthetic corpora remain qualitatively similar when varying the parameter of the degree of structure, $c$.

\begin{figure*}[h]
	\begin{center}
		\centerline{\includegraphics[width=1.8 \columnwidth]{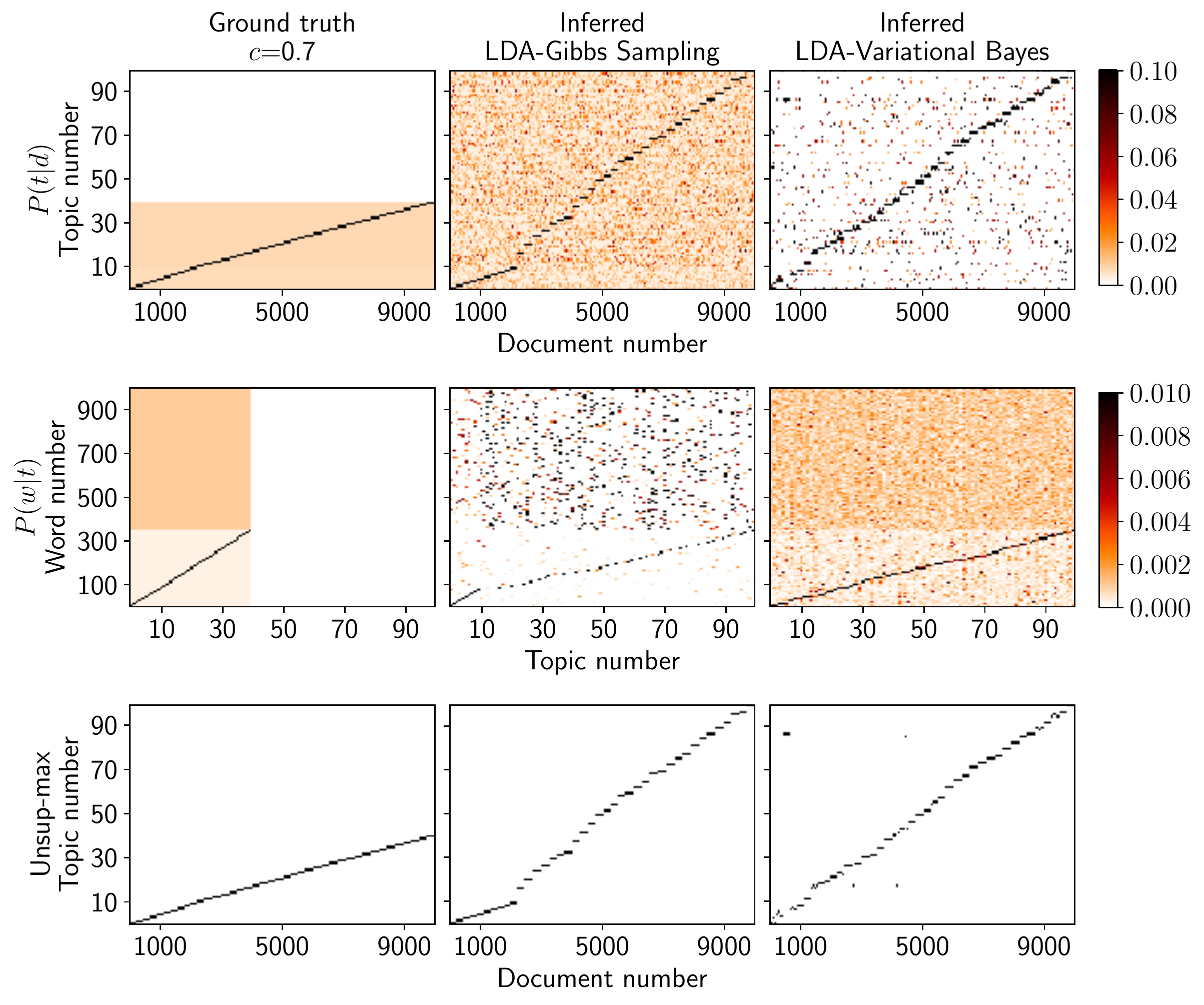}}
		\caption{
			\textbf{Different LDA models lead to qualitatively different solutions in the case of stopwords.}
			Comparison of the topic-document distribution (top row), $P(t|d)$ , word-topic distribution (middle row), $P(w|t)$, and predicted topic in unsupervised document classification (bottom row), $\underset{t}{\arg \max}P(t|d) $ for three cases:
			Ground truth as planted in the synthetic corpus (left column), inferred from Gibbs Sampling LDA (middle column), and inferred from Variational Bayes LDA (right column).
			Same parameters as in Fig.~5 
			(E, F) setting the fraction of stopwords $P_s = 0.65$ and the degree of structure $c=0.7$.
		}
		\label{fig_changPS_compare}
	\end{center}
\end{figure*}

In Fig.~\ref{fig_changPS_compare} we show differences between the planted and inferred topic distributions $P(t|d)$ and $P(w|t)$ for synthetic corpora in the case of stopwords.

\begin{figure*}[h]
	\begin{center}
		\centerline{\includegraphics[width=1.6 \columnwidth]{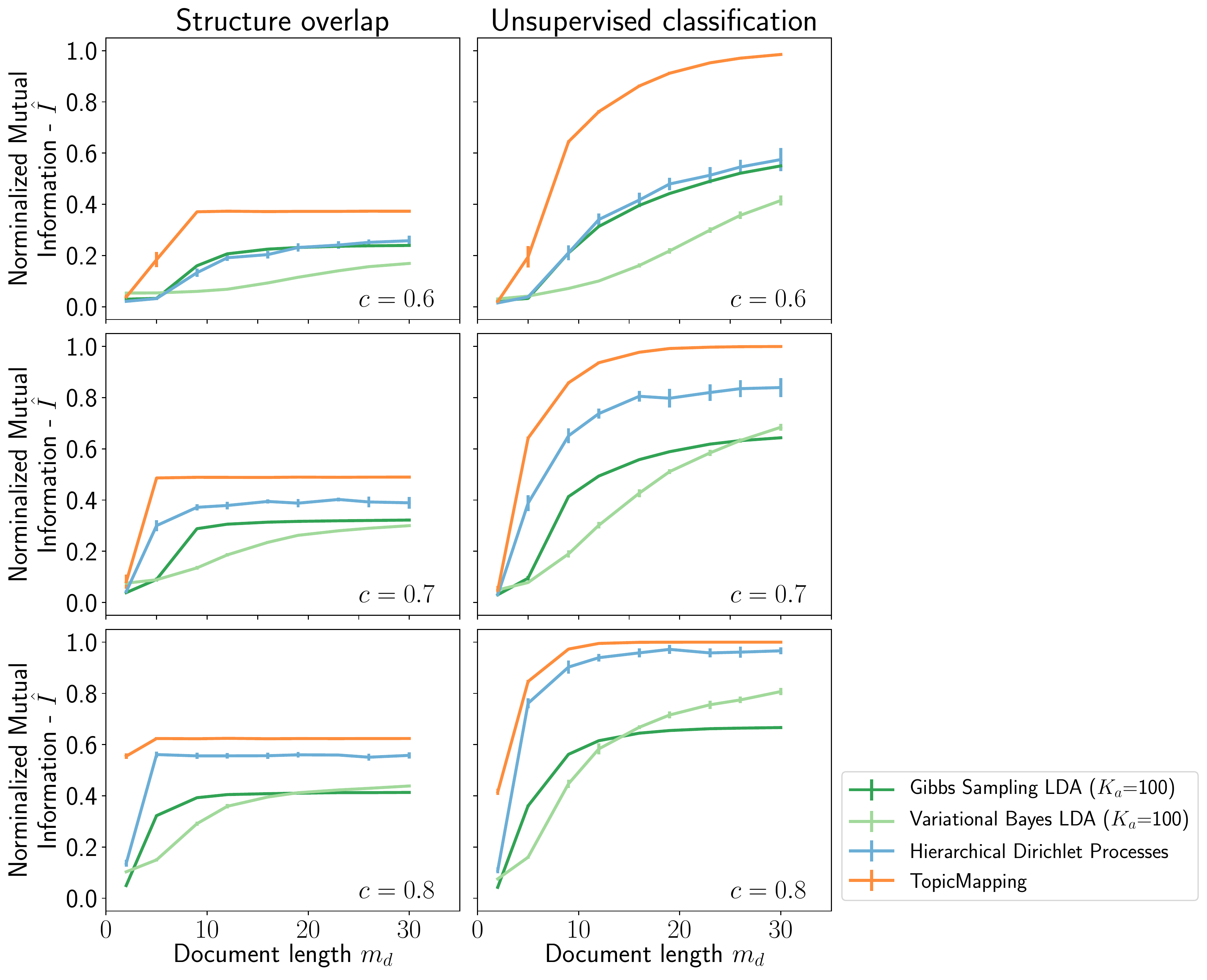}}
		\caption{
			\textbf{Varying the degree of structure does not effect results on document length dependency in synthetic corpora.}
			Normalized mutual information, $\hat{I}$, as measured by structure overlap (left column) and unsupervised document classification (right column) as in Fig.~5 
			(B, C) varying the degree of structure:
			$c=0.6$ (top row), $c=0.7$ (middle row), and $c=0.8$ (bottom row).
		}
		\label{fig_changeC_changedocLength_compare}
	\end{center}
\end{figure*}

In Fig.~\ref{fig_changeC_changedocLength_compare} we show that the results from Fig.~5 
(E, F) investigating the effect of document length on the performance of topic models in synthetic corpora remain qualitatively similar when varying the parameter of the degree of structure, $c$.

\clearpage


\bibliography{aistats2019_topic_model,aistats2019_topic_model_package}